\documentclass[11pt]{article}

\usepackage[preprint]{acl}

\usepackage{times}
\usepackage{latexsym}

\usepackage[T1]{fontenc}

\usepackage[utf8]{inputenc}

\usepackage{microtype}

\usepackage{inconsolata}

\usepackage{graphicx}
\usepackage{hyperref}
\usepackage{amsmath}
\usepackage{booktabs}
\usepackage{enumitem}
\usepackage{amssymb}
\usepackage{array}
\usepackage{tcolorbox}
\usepackage{multirow}
\usepackage{xcolor}
\usepackage{pifont}
\usepackage{enumitem}
\usepackage[table]{xcolor}
\usepackage{tabularx}
\usepackage{float}
\usepackage{makecell}
\usepackage{xcolor}
\usepackage{subcaption}
\usepackage{mathtools}

\definecolor{descblue}{RGB}{40, 90, 160}
\definecolor{boundred}{RGB}{180, 50, 50}

\newcommand{\red}[1]{\textcolor{black}{#1}}

\newcommand{\ours}{\textsc{MusTBench}}
\newcommand{\oursmodel}{\textsc{MusT}} 

\newcommand{\tightparagraph}[1]{\vspace{2.5pt}\noindent\textbf{#1}}

\title{\ours{}: Benchmarking and Advancing \\ Temporal Grounding in Music LLMs}

\author{
\\
  \textbf{Daeyong Kwon\textsuperscript{1,2}\thanks{Work done during an internship at Sony Group Corporation.}},
  \textbf{Qiyu Wu\textsuperscript{2}\thanks{Corresponding author: Qiyu Wu, \texttt{qiyu.wu@sony.com}.}},
  \textbf{Shinobu Kuriya\textsuperscript{2}},
  \textbf{Junghyun Koo\textsuperscript{3}},
  \textbf{Shuyang Cui\textsuperscript{2}},
  \\
  \textbf{Zhi Zhong\textsuperscript{2}},
  \textbf{Wei-Hsiang Liao\textsuperscript{3}},
  \textbf{Hiromi Wakaki\textsuperscript{2}},
  \textbf{Yuki Mitsufuji\textsuperscript{2,3}}
  \vspace{4pt}\\
  \textsuperscript{1}Seoul National University
  \quad
  \textsuperscript{2}Sony Group Corporation
  \\
  \textsuperscript{3}Sony Corporate Technology Center America, Inc.
  \\
  \fontsize{10pt}{10pt}\selectfont\textsuperscript{1}\texttt{daeyongkwon@snu.ac.kr}
  \quad
  \textsuperscript{2,3}\texttt{\{qiyu.wu, hiromi.wakaki, yuhki.mitsufuji\}@sony.com}
  \\\vspace{4pt}
    \href{https://daeyongkwon98.github.io/mustbench/}{
      \textcolor{magenta}{\texttt{https://daeyongkwon98.github.io/mustbench/}}
    }
}

\begin{document}
\maketitle

\begin{abstract}
Recent Large Audio-Language Models (LALMs) have demonstrated promising abilities in understanding musical content. However, whether their responses are grounded in the correct temporal regions of the audio remains underexplored. This limitation is particularly critical for music understanding, where key information often occurs as temporally localized events, such as instrument entries and rhythmic transitions.
To address this gap, we introduce \ours{}, a reliable benchmark with stem-wise data and music-expert validation designed to evaluate temporal grounding in LALMs through five temporally grounded question-answering tasks. To further improve temporal grounding in existing models, we propose \oursmodel{}, a novel four-stage temporal optimization recipe spanning music encoder adaptation, LLM adaptation, LLM supervised fine-tuning, and RL-based optimization.
Experiments on \ours{} show that existing LALMs face challenges in precise temporal grounding, while \oursmodel{} consistently improves performance
across different LALM backbones and scales. These results highlight temporal grounding as an important challenge in current LALMs and provide a benchmark and training recipe that may support future research on temporally grounded music understanding.\footnote{Code and benchmark data are publicly available.}
\end{abstract}

\section{Introduction}

\begin{figure*}[t]
  \centering
  \includegraphics[width=\textwidth]{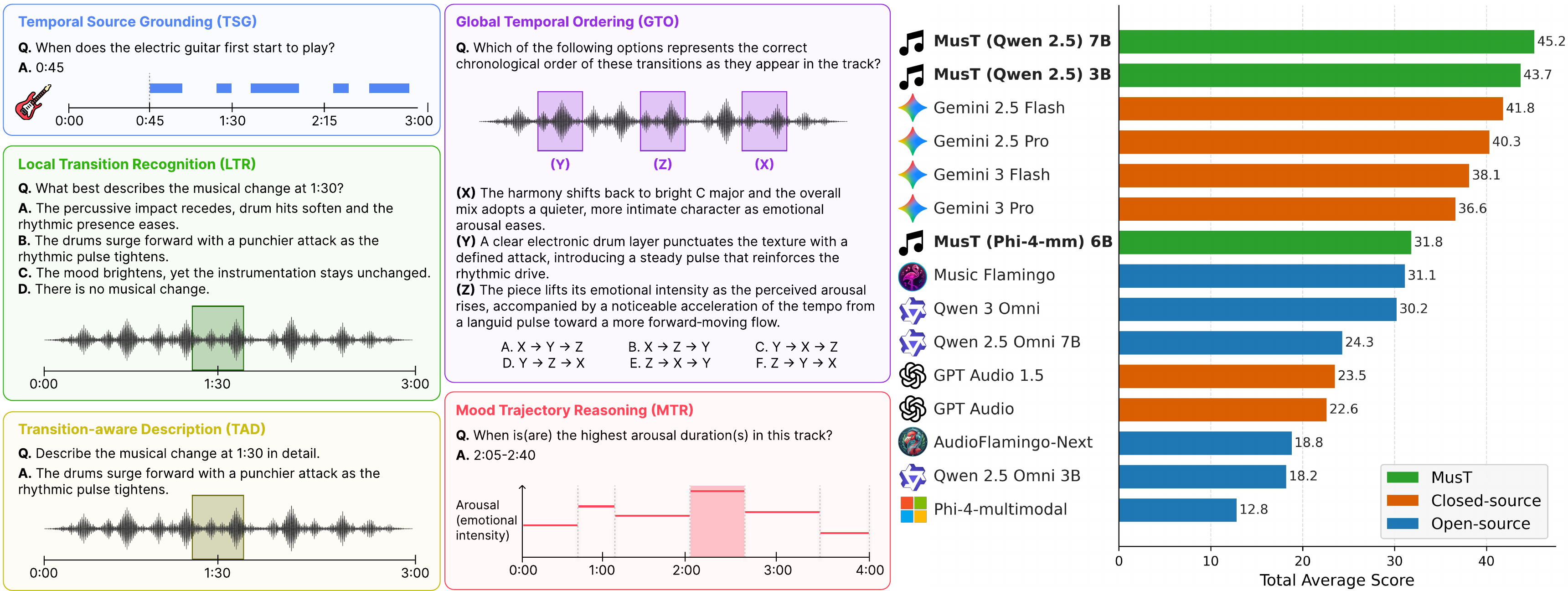}
  \caption{
  \textbf{(Left)} \ours{} examples illustrating five types of temporal grounding music questions. \red{\textbf{(Right)} Performance on \ours{} comparing baselines with \oursmodel{}, showing consistent gains from our approach.}
  }
  \vspace{-1em}
  \label{fig:first_figure}
\end{figure*}

Music understanding has long been a challenging problem due to the complex attributes in musical audio.
Recent advances~\cite{tang2023salmonn, chu2024qwen2} in Large Audio-Language Models (LALMs) have extended their capabilities to non-speech domains such as music.
State-of-the-art models such as Qwen3-Omni~\citep{xu2025qwen3} and Music Flamingo~\citep{ghosh2025music} can generate descriptions of musical tracks with rich attributes, suggesting that LALMs are increasingly capable of understanding complex musical content.

However, detailed music description does not necessarily indicate temporal grounding.
Temporal grounding is particularly critical for music understanding, as key information often occurs through temporally localized or evolving events, such as instrument entries and shifts in timbre, harmony, or tempo~\cite{paulus2010state}.
Since LALMs are known to sometimes hallucinate~\cite{cheng2026aha}, a plausible description such as \textit{``the track evolves with the entry of a guitar''} is not sufficient evidence that the model is grounded in the audio. 
Such a claim remains difficult to verify unless the model can also identify the temporal region where the guitar enters.
We refer to this capability as \textit{music temporal grounding}: the ability to associate a textual claim about musical content with the specific time point or interval in the audio where the claim is acoustically supported.
However, existing music benchmarks mainly evaluate clip-level captioning~\cite{agostinelli2023musiclm} or general audio question answering (QA)~\cite{weck2024muchomusic}, leaving temporal grounding largely underexplored.

\red{To address this gap, we introduce \textbf{Mus}ic \textbf{T}emporal \textbf{Bench}mark (\textbf{\ours{}}), a reliable benchmark with stem-wise data and music-expert validation for evaluating temporal grounding in LALMs.}
\ours{} comprises five temporally grounded QA tasks: temporal source grounding (TSG), local transition recognition (LTR), transition-aware description (TAD), global temporal ordering (GTO), and mood trajectory reasoning (MTR). 
Together, these tasks assess a model's ability to ground musical events to the specific moments or intervals at which they occur. Figure~\ref{fig:first_figure} presents examples from each task category and provides a preview of model performance.
Our evaluation shows that existing models remain limited in temporal grounding. 
For example, in TSG, the most basic grounding task, a model is given a full music track and asked to predict when an instrument or vocal first enters or finally exits. 
Even on this straightforward task, current LALMs exhibit systematic failures. 
They struggle particularly with identifying offsets, often collapse their predictions to coarse temporal anchors such as 60s or 120s, and sometimes generate timestamps that fall outside the valid duration of the audio. 
Together with the broader diagnostic on \ours{}, these observations suggest that current models often rely on rough temporal priors rather than precisely grounding their responses in the input audio, indicating a lack of basic temporal perception ability.

To further improve temporal grounding in existing models, we propose \oursmodel{}, a four-stage temporal optimization recipe spanning music encoder adaptation, LLM adaptation with timestamped music captions, supervised temporal QA fine-tuning, and RL-based optimization.  This recipe equips the model with transition-aware temporal representations, adapts the LLM to timestamped music understanding, and directly optimizes timestamp- and interval-level grounding ability. 
\red{Experiments show that \oursmodel{} improves performance across different LALM backbones.}

Our contributions are summarized as follows: \textbf{(1)} We identify temporal grounding as a key missing capability in current LALMs for music understanding; \red{\textbf{(2)} We introduce \ours{}, a reliable benchmark with stem-wise data and music-expert validation for evaluating temporal grounding in full-length music through five temporally grounded QA tasks; and \textbf{(3)} We propose \oursmodel{}, a four-stage temporal optimization recipe that improves temporal grounding across different LALM backbones.} Together, these contributions provide a challenging benchmark and a practical training recipe for advancing temporal grounding in music understanding.

\section{Related Work}

\paragraph{Music LLMs.}
Recent LALMs have expanded music understanding from fixed-label prediction to captioning, question answering, and instruction following~\cite{agostinelli2023musiclm, deng2024musilingo}. 
Music-oriented LLMs such as MU-LLaMA~\citep{liu2024music} and MuMu-LLaMA~\citep{liu2025mumu} align music representations with language models, while general audio-language models such as the Qwen-Omni series~\citep{xu2025qwen2, xu2025qwen3}, and Music Flamingo~\citep{ghosh2025music} have shown strong performance on music reasoning tasks.
These models can generate descriptions of musical attributes such as mood, genre, and tempo, while their temporal grounding ability remains underexplored.

\tightparagraph{Music Understanding Benchmarks.}
Existing audio-based music benchmarks primarily evaluate global music understanding. 
Datasets such as MusicQA~\citep{liu2024music}, MusicInstruct~\citep{deng2024musilingo}, MusicCaps~\citep{agostinelli2023musiclm}, MagnaTagATune~\citep{law2009evaluation}, and MuChoMusic~\citep{weck2024muchomusic} test whether models can recognize musical content or answer questions about the overall audio. 
While useful for evaluating music perception and language generation, these benchmarks rarely require models to ground an answer to a specific timestamp or temporal interval.

\tightparagraph{Temporal Grounding.}
Temporal grounding has been studied in video, audio-visual, and general audio understanding, where models localize moments or sound events from natural language queries~\citep{li2022learning, chowdhury2024meerkat, xu2021text}. 
Recent studies show that audio-language models can exhibit temporal biases, including hallucinated events and incorrect chronological ordering~\citep{yao2025not}, and timestamp-aware audio captioning can improve temporal alignment~\citep{kumar2026tac}. 
\red{More recent approaches improve fine-grained temporal grounding through explicit temporal representations and post-training~\citep{sun2026spotsound, shi2026towards}.}
However, music poses a distinct challenge, as meaningful changes often emerge through evolving instrumentation, rhythm, harmony, and emotion, rather than appearing as isolated discrete sound events.
In this work, we fill this gap by evaluating whether LALMs can ground such music-specific changes to concrete moments in the audio, and also provide a practical training recipe to improve existing models.

\begin{figure*}[t]
  \centering
  \includegraphics[width=\textwidth]{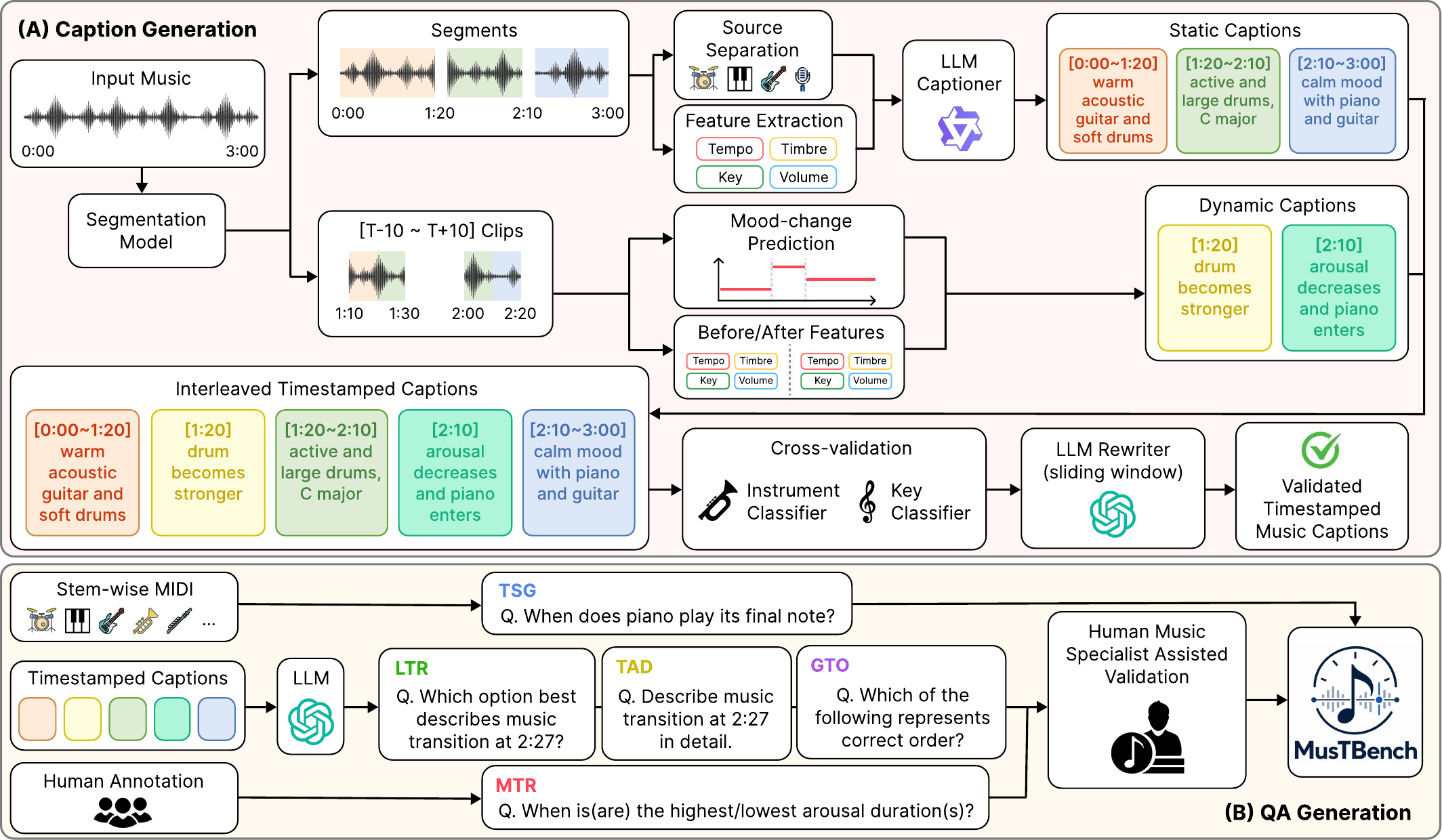}
\caption{
Overview of the \ours{} construction pipeline. 
\textbf{(A)} Timestamped music captions are generated by segmentation, mood-change modeling, feature-grounded captioning, and cross-validation and rewriting.
\textbf{(B)} QA pairs are generated from stem-wise MIDI annotations, timestamped music captions, and human annotations. 
\red{Generated QA pairs are validated with assistance from human music experts to produce the final benchmark.}
}
\vspace{-1em}
  \label{fig:caption_generation_pipeline}
\end{figure*}

\section{\ours{}}\label{sec:tempoground}

We introduce \ours{}: a benchmark for evaluating temporal grounding in music understanding. The construction of \ours{} consists of (1) building a timestamped music caption dataset, and (2) generating five types of QA tasks for different aspects of temporal grounding.
Figure~\ref{fig:caption_generation_pipeline} provides an overview of the full data generation pipeline.

\paragraph{Data Source and Usage.}
\hypertarget{paper:data-usage-lines}{}%
\red{\ours{} is constructed from MTG-Jamendo~\citep{bogdanov2019mtg} and Slakh2100~\citep{manilow2019cutting} test tracks. These tracks are used exclusively for benchmark construction and evaluation. Training examples are constructed from disjoint Slakh2100 train tracks and an internal music dataset. No song used for training is included in the final evaluation benchmark. (Table~\ref{tab:appendix_train_statistics})}

\subsection{Timestamped Music Caption Generation}
\label{sec:timestamped_music_caption}

\red{We first build timestamped captions from full-track audios, which serve as intermediate annotations for identifying musical states and transitions.}
As shown in Figure~\ref{fig:caption_generation_pipeline}-A, the process consists of four steps:
\begin{enumerate} [leftmargin=*, noitemsep, topsep=4pt]
    \item \textbf{Segmentation.}
    We apply a structural music segmentation model~\citep{hao2025songformer} to obtain segment boundaries, which serve as candidate transition timestamps.

    \item \textbf{Mood-change modeling.}
    \red{We use a trained mood predictor to estimate changes in emotional intensity around segment boundaries.}

    \item \textbf{Feature-grounded captioning.}
    We extract diverse musical features and use them to generate segment-level static captions and boundary-level dynamic captions.

    \item \textbf{Cross-validation and rewriting.}
    We verify instrument and key descriptions using additional music classifiers, then rewrite neighboring captions to improve coherence while preserving timestamps and musical facts.
\end{enumerate}
The resulting timestamped captions provide temporally aligned musical evidence, used for later timestamped pretraining and QA generation. Implementation details are provided in \S\ref{sec:appendix_timestamped_caption_generation}.

\subsection{\ours{} QA Generation}\label{sec:qa_dataset}


\begin{table}[t]
\centering
\small
\setlength{\tabcolsep}{3.5pt}
\resizebox{.95\columnwidth}{!}{%
\begin{tabular}{@{} l | r | r r r r r @{}}
\toprule
\textbf{Statistic} 
& \textbf{Overall} 
& \textbf{TSG} 
& \textbf{LTR} 
& \textbf{TAD} 
& \textbf{GTO} 
& \textbf{MTR} \\
\midrule
\# QA pairs
& 1,264 & 400 & 208 & 208 & 198 & 250 \\
\# Unique Songs
& 875 & 246 & 208 & 208 & 198 & 250 \\
Avg. duration
& 3m 32s & 3m 51s & 3m 31s & 3m 31s & 3m 26s & 3m 18s \\
\bottomrule
\end{tabular}%
}
\caption{\red{Statistics of \ours{}. The overall song count is computed after merging overlaps across task types. Audio duration is computed over unique songs.}}
\vspace{-10pt}
\label{tab:tempoground_statistics}
\end{table}
To comprehensively evaluate temporal grounding ability, we propose five types of task. \red{We use gpt-oss-120b~\citep{agarwal2025gpt} to generate QAs.}

\tightparagraph{Temporal Source Grounding.}
TSG measures the most fundamental temporal grounding ability by asking models to identify when a specific sound source begins or ends.
\red{For instrument questions, we extract onset and offset times from MIDI-aligned instrument tracks in Slakh2100, using disjoint tracks for training and benchmark evaluation.
For vocal questions, we use an internal dataset for training and MTG-Jamendo for evaluation\footnote{Our use of the MTG-Jamendo dataset is limited solely to evaluation; it is not used for training or developing our models.}, applying a source separator~\cite{rouard2023hybrid} and volume-based filtering to determine vocal onset and offset times.}
For both question types, we retain only examples where the target source is sufficiently audible and ask only about its first or final audible occurrence, making the queried source perceptually identifiable and the ground truth temporally unambiguous.

\tightparagraph{Local Transition Recognition.}
LTR evaluates whether a model can identify the description corresponding to a specified transition timestamp. Given the full audio and a timestamp, the model must select the description that best matches the musical transition occurring at that moment from multiple choices.
To make this task more challenging, we primarily use real transition descriptions from the same song but different timestamps as distractors, to prevent the shortcut where the model relies only on the overall musical context.

\tightparagraph{Transition-Aware Description.}
TAD evaluates open-ended transition understanding. Given the full audio and a transition timestamp, the model is asked to describe the musical change occurring around that moment. 
This removes answer-space constraints and tests whether the model can generate a correct temporally grounded description without relying on predefined options.

\tightparagraph{Global Temporal Ordering.}
GTO evaluates relative temporal reasoning over multiple transition events. We sample three transition descriptions from the same track and present them in random order. The model is then asked to determine their correct chronological order within the song, out of the six possible permutations.
Unlike LTR and TAD which focus on local transitions, this type assesses whether the model can reason over the global temporal structure of multiple events distributed throughout the track.

\tightparagraph{Mood Trajectory Reasoning.}
MTR evaluates whether models can localize the time intervals where the music reaches its highest or lowest emotional intensity.
For training, we construct temporal annotations using a trained mood-change predictor.
\red{Given transition boundaries, we estimate the arousal trajectory by cumulatively summing predicted mood changes and identify segments with the highest or lowest values.}
We filter out overly short segments and cases with insufficient separation between high- and low-arousal regions, while allowing multiple answer spans when segments have similar arousal levels.
\red{For benchmark evaluation, music experts directly listen to the audio and annotate the highest or lowest arousal intervals.}

Finally, we apply task-specific validation procedures to obtain 1,264 high-quality QA pairs. \red{TSG instances are validated using stem-wise data, while the remaining tasks are validated by human music experts. Details of the validation procedure are provided in \S\ref{sec:appendix_human_validation}. Table~\ref{tab:tempoground_statistics} summarizes the statistics of \ours{}, with additional details provided in Tables~\ref{tab:appendix_train_statistics}--\ref{tab:benchmark_tag_analysis}.}



\subsection{Evaluation Metrics}
\label{sec:evaluation_metric}
Since \ours{} evaluates various abilities, we use task-specific evaluation metrics to measure performance. For LTR and GTO, we report accuracy as they are multiple-choice tasks.

{
\setlength{\abovedisplayskip}{2pt}
\setlength{\belowdisplayskip}{2pt}
\setlength{\abovedisplayshortskip}{2pt}
\setlength{\belowdisplayshortskip}{2pt}

\tightparagraph{Temporal Source Grounding.}
For TSG, the model predicts a single timestamp. 
We evaluate whether the prediction falls within a tolerance threshold $T$ of the ground-truth timestamp:
\vspace{3pt}
\[
\text{Hit@T} =
\frac{1}{N}
\sum_{i=1}^{N}
\mathbb{I}
\left(
\left| t_{\mathrm{pred}}^{(i)} - t_{\mathrm{gt}}^{(i)} \right| \leq T
\right).
\]

\tightparagraph{Transition-Aware Description.}
For TAD, the model generates a free-form textual answer. 
We report METEOR~\cite{banerjee2005meteor} for lexical overlap with reference description and CLAP Score~\cite{laionclap2023} for audio-text alignment. 
The CLAP Score is computed as the cosine similarity between the audio embedding of a 20-second transition clip and the text embedding of the generated description using \textit{"laion/clap-htsat-fused"}.

\tightparagraph{Mood Trajectory Reasoning.}
For MTR, a question may have multiple valid temporal durations.
Let $\mathcal{P}$ and $\mathcal{G}$ denote the predicted and ground-truth duration sets, and let $U(\cdot)$ denote the union of durations in a set.
We report temporal IoU and F1, which measure the overlap between predicted and ground-truth temporal regions:
\vspace{3pt}
\[
\begin{aligned}
\text{IoU} &=
\frac{
\left| U(\mathcal{P}) \cap U(\mathcal{G}) \right|
}{
\left| U(\mathcal{P}) \cup U(\mathcal{G}) \right|
},
&
\text{F1} &=
\frac{
2 \left| U(\mathcal{P}) \cap U(\mathcal{G}) \right|
}{
\left| U(\mathcal{P}) \right| + \left| U(\mathcal{G}) \right|
}
\end{aligned}
\]

\subsection{Performance of Existing Models}
\label{sec:baseline_performance}

We evaluate representative open- and closed-source LALMs, as listed in \S\ref{app:experimental_details}. 
Table~\ref{tab:main_results} shows a clear gap between recognizing temporal information and precisely localizing it. 
Open-source models obtain relatively strong results on the multiple-choice tasks, LTR and GTO, where they only need to select among predefined options. 
However, their performance drops substantially on tasks that require direct temporal prediction. 
For example, several open-source models achieve competitive GTO accuracy, but their MTR scores remain consistently low, indicating difficulty in estimating continuous temporal boundaries. 
Closed-source models generally perform better on these localization-heavy tasks, but their results are still far from saturated. 
These findings suggest that current LALMs can often recognize temporally relevant musical changes, but precise temporal boundary estimation remains a major unresolved challenge.
In addition, the results show that music temporal grounding is not simply determined by general model capability.
Within the Gemini family, Flash variants outperform their Pro counterparts, and Gemini 2.5 obtain higher overall scores than Gemini 3.
This suggests that temporal grounding is a specialized capability that depends on fine-grained audio-time alignment, rather than a direct byproduct of stronger reasoning ability.

\tightparagraph{Qualitative Analysis on TSG.}
\begin{figure}[t]
  \centering
  \includegraphics[width=.9\columnwidth]{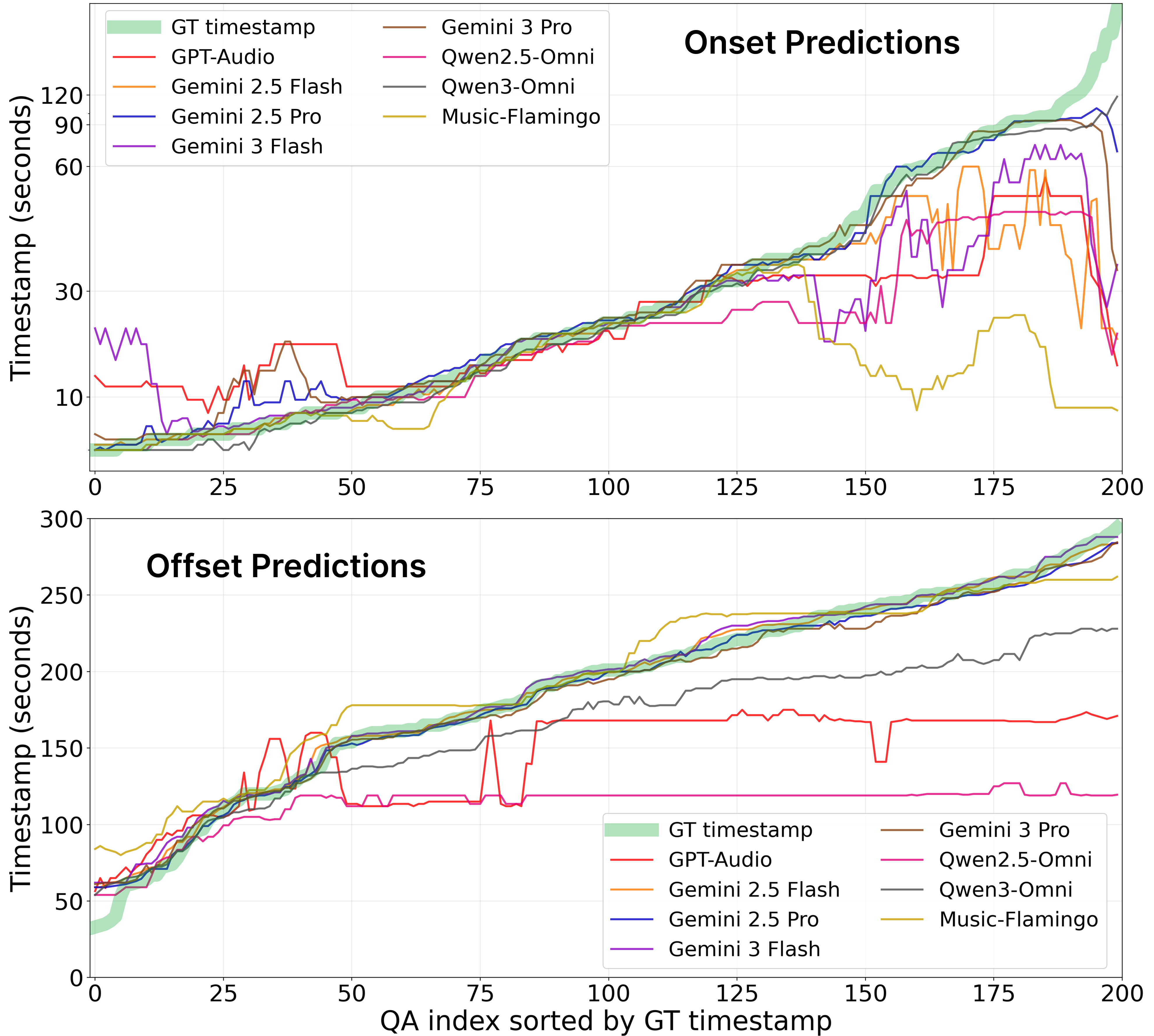}
    \caption{
    TSG predictions for onset and offset localization.
    To show overall trends, each line is smoothed using a sliding-window mean with a window size of 20.
    }
  \label{fig:onset_offset_lines}
  \vspace{-1em}
\end{figure}
To better understand behavior, we analyze TSG predictions in Figure~\ref{fig:onset_offset_lines}.
Overall, most models more reliably identify when a source first appears than when it disappears.
For onset prediction, models are most accurate when the ground-truth timestamp falls around 10--30 seconds, while accuracy decreases for both early and later entries.
Offset prediction shows heavier temporal instability.
Models perform reasonably well within roughly the first 120 seconds, but predictions become increasingly noisy afterward, often collapsing to coarse duration anchors such as 120 and 180 seconds.
These patterns suggest that some models rely on temporal priors rather than accurately grounding offsets in the input audio.
Among the evaluated models, the Gemini series shows the strongest performance, while Qwen3 Omni remains competitive on onset prediction.
We also observe out-of-range predictions beyond the input audio duration, indicating that current LALMs do not always respect input-specific temporal constraints.
These limitations motivate our transition-aware dual-encoder model for fine-grained music temporal grounding.
Full prediction details are provided in Figure~\ref{fig:eight_figures}.

\section{\oursmodel{} Training}

\begin{table*}[t]
\centering
\small
\setlength{\tabcolsep}{2.5pt}
\resizebox{\textwidth}{!}{%
\begin{tabular}{llcccccccccccc}
\toprule
& 
& \multicolumn{3}{c}{\textbf{TSG}}
& \textbf{LTR}
& \multicolumn{3}{c}{\textbf{TAD}}
& \textbf{GTO}
& \multicolumn{3}{c}{\textbf{MTR}}
& \textbf{Total} \\
\cmidrule(lr){3-5}
\cmidrule(lr){6-6}
\cmidrule(lr){7-9}
\cmidrule(lr){10-10}
\cmidrule(lr){11-13}
\cmidrule(lr){14-14}
\textbf{Model}
& \textbf{Size}
& \makecell{\textbf{Onset} \\ \textbf{Hit@3s}}
& \makecell{\textbf{Offset} \\ \textbf{Hit@3s}}
& \textbf{Avg.}
& \textbf{Acc.}
& \textbf{METEOR}
& \makecell{\textbf{CLAP} \\ \textbf{Score}}
& \textbf{Avg.}
& \textbf{Acc.}
& \makecell{\textbf{Temporal} \\ \textbf{IoU}}
& \makecell{\textbf{Temporal} \\ \textbf{F1}}
& \textbf{Avg.}
& \textbf{Avg.} \\
\midrule

\multicolumn{14}{l}{\textcolor{gray}{\textit{Closed-source Models}}} \\
Gemini 2.5 Flash 
& -- & 60.0 & 71.5 & 65.8 & 56.7 & 11.2 & 34.5 & 22.9 & 42.4 & 24.8 & 33.1 & 29.0 & 41.8 \\
Gemini 2.5 Pro   
& -- & 57.5 & 57.0 & 57.3 & 62.5 & 10.0 & 38.1 & 24.1 & 46.0 & 22.4 & 28.6 & 25.5 & 40.3 \\
Gemini 3 Flash   
& -- & 55.0 & 42.5 & 48.8 & 51.4 & 10.7 & 37.9 & 24.3 & 47.0 & 25.7 & 34.2 & 30.0 & 38.1 \\
Gemini 3 Pro     
& -- & 60.0 & 44.5 & 52.3 & 54.3 & 6.5 & 33.0 & 19.8 & 27.3 & 29.2 & 37.6 & 33.4 & 36.6 \\
GPT Audio        
& -- & 21.5 & 1.5  & 11.5 & 42.3 & 12.0 & 33.0 & 22.5 & 36.9 & 13.8 & 19.6 & 16.7 & 22.6 \\
GPT Audio 1.5    
& -- & 14.0 & 3.5  & 8.8 & 51.0 & 12.4 & 34.8 & 23.6 & 37.9 & 13.7 & 20.3 & 17.0 & 23.5 \\

\midrule
\multicolumn{14}{l}{\textcolor{gray}{\textit{Open-source Models}}} \\
Phi-4-mm             
& 6B & 14.5 & 0.0 & 7.3 & 28.4 & 9.6 & 23.0 & 16.3 & 13.1 & 5.0 & 8.8 & 6.9 & 12.8 \\
AF-Next              
& 8B & 17.0 & 3.0 & 10.0 & 44.7 & 9.0 & 29.4 & 19.2 & 41.4 & 2.2 & 3.4 & 2.8 & 18.8 \\
Music Flamingo       
& 8B & 53.0 & 24.0 & 38.5 & 56.3 & 13.4 & 33.4 & 23.4 & 41.4 & 10.1 & 17.1 & 13.6 & 31.1 \\
Qwen 2.5 Omni        
& 3B & 7.5 & 2.0 & 4.8 & 32.7 & 11.0 & 33.6 & 22.3 & 37.4 & 8.2 & 12.8 & 10.5 & 18.2 \\
Qwen 2.5 Omni        
& 7B & 39.0 & 3.5 & 21.3 & 45.7 & 9.2 & 33.7 & 21.5 & 46.5 & 6.4 & 10.6 & 8.5 & 24.3 \\
Qwen 3 Omni          
& 30B-A3B & 62.5 & 9.5 & 36.0 & 53.4 & 7.1 & 24.9 & 16.0 & 63.6 & 8.8 & 12.0 & 10.4 & 30.2 \\

\midrule
\textbf{\oursmodel{} \scriptsize(Phi-4-mm)}
& 6B
& 20.5 {\scriptsize(+6.0)}
& 34.0 {\scriptsize(+34.0)}
& 27.3 {\scriptsize(+20.0)}
& 66.8 {\scriptsize(+38.4)}
& 19.9 {\scriptsize(+10.3)}
& 24.4 {\scriptsize(+1.4)}
& 22.2 {\scriptsize(+5.9)}
& 51.5 {\scriptsize(+38.4)}
& 14.3 {\scriptsize(+9.3)}
& 22.8 {\scriptsize(+14.0)}
& 18.6 {\scriptsize(+11.7)}
& 31.8 {\scriptsize(+19.0)} \\

\textbf{\oursmodel{} \scriptsize(Qwen 2.5 Omni)}
& 3B 
& 60.5 {\scriptsize(+53.0)}
& 64.0 {\scriptsize(+62.0)}
& 62.3 {\scriptsize(+57.5)}
& 70.2 {\scriptsize(+37.5)}
& 22.8 {\scriptsize(+11.8)}
& 30.6 {\scriptsize(-3.0)}
& 26.7 {\scriptsize(+4.4)}
& 57.6 {\scriptsize(+20.2)}
& 17.5 {\scriptsize(+9.3)}
& 26.5 {\scriptsize(+13.7)}
& 22.0 {\scriptsize(+11.5)}
& 43.7 {\scriptsize(+25.5)} \\

\textbf{\oursmodel{} \scriptsize(Qwen 2.5 Omni)}
& 7B
& 65.5 {\scriptsize(+26.5)}
& 65.0 {\scriptsize(+61.5)}
& 65.3 {\scriptsize(+44.0)}
& 74.0 {\scriptsize(+28.3)}
& 22.4 {\scriptsize(+13.2)}
& 30.3 {\scriptsize(-3.4)}
& 26.4 {\scriptsize(+4.9)}
& 61.1 {\scriptsize(+14.6)}
& 17.5 {\scriptsize(+11.1)}
& 25.6 {\scriptsize(+15.0)}
& 21.6 {\scriptsize(+13.1)}
& 45.2 {\scriptsize(+20.9)} \\
\bottomrule
\end{tabular}
}
\caption{\red{
Evaluation results on \ours{}. All metrics are in a 0–100 scale. The "Avg." columns under TSG, TAD, and MTR denote the mean of their corresponding sub-metrics, and "Total Avg." indicates the overall average across eight metrics. Parenthesized numbers represent absolute gains over the corresponding base model.}
}
\vspace{-13pt}
\label{tab:main_results}
\end{table*}
\label{sec:training_pipeline}

The analysis on \ours{} demonstrates that existing open-source models are still limited in temporally grounded music understanding. To further improve them, we propose \oursmodel{}, a four-stage training recipe as illustrated in Figure~\ref{fig:model_architecture}. 
\red{We adopt Qwen2.5-Omni~\citep{xu2025qwen2} and Phi-4-multimodal~\citep{abouelenin2025phi} as the backbone to examine its applicability across different architectures.}

\tightparagraph{Stage 1: Transition-Aware Encoder Pretraining.}
As we discussed above, existing LALMs can lack basic temporal perception. Hence, we first adapt a MERT encoder~\cite{li2024mert} with LoRA~\citep{hu2021lora} to better capture temporally meaningful musical changes.
The encoder is trained with two temporal objectives: transition probability prediction and mood-change prediction.
For transition probability prediction, we use structural segment boundaries to construct Gaussian-smoothed boundary targets.
Given the predicted transition logit $z_t$ at frame $t$, the predicted probability $p_t=\sigma(z_t)$, and the Gaussian-smoothed target $y_t$, we optimize a BCE-Dice loss:
\vspace{3pt}
\[
\mathcal{L}_{\mathrm{trans}}
=
\mathrm{BCE}(z,y)
+
1 -
\frac{
2\sum_t p_t y_t + s
}{
\sum_t p_t + \sum_t y_t + s
}.
\]
\vspace{3pt}

Here, $s$ is a smoothing constant.
The BCE term provides frame-level supervision, while the Dice term encourages overlap between predicted transition probabilities and sparse boundary regions.
For mood-change prediction, we train the encoder to estimate changes in emotional intensity around transition boundaries using the annotations in \S\ref{sec:timestamped_music_caption}.
Given predicted and gold mood-change scores $\hat{a}$ and $a$, we use the concordance correlation coefficient loss (CCC loss):
\vspace{2pt}
\[
\mathcal{L}_{\mathrm{mood}}
=
1 -
\frac{
2\,\mathrm{cov}(\hat{a}, a)
}{
\sigma_{\hat{a}}^2
+
\sigma_a^2
+
(\mu_{\hat{a}}-\mu_a)^2
+
\epsilon
}.
\]

The encoder is optimized using the sum of the transition and mood-change losses.
The proposed \oursmodel{} encoder provides fine-grained temporal representations for the following stages.
Details are described in \S\ref{app:encoder_pretraining}.

\begin{figure*}[t]
  \centering
  \includegraphics[width=\textwidth]{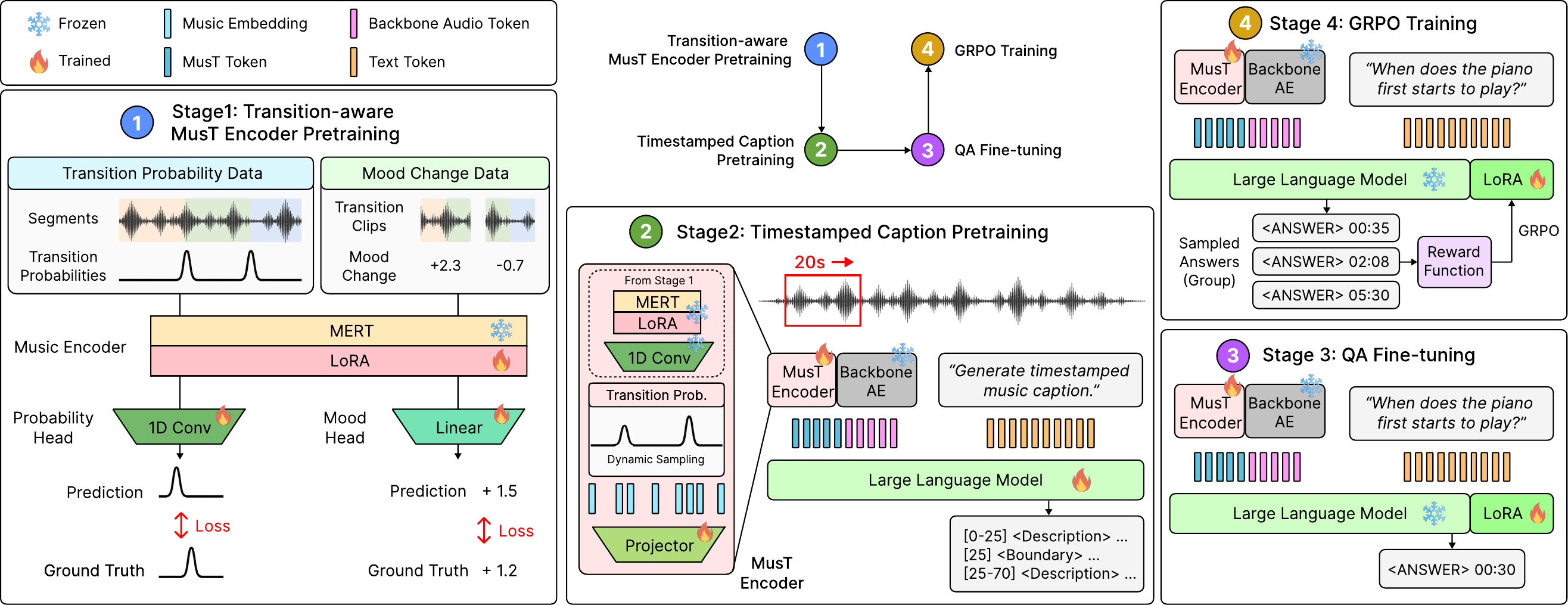}
    \caption{\red{
    Overview of the proposed four-stage training pipeline (\oursmodel{}) and model architecture.}
}
\vspace{-1em}
  \label{fig:model_architecture}
\end{figure*}

\tightparagraph{Stage 2: Timestamped Caption Pretraining.}
Inspired by~\citet{tang2023salmonn}, we train the dual-encoder LLM on timestamped music captioning for modality alignment.
\oursmodel{} encoder provides fine-grained acoustic and temporal representations, while the backbone's frozen audio encoder provides semantic representations.
A learnable projector maps \oursmodel{} encoder outputs into the LLM embedding space, and sinusoidal time embeddings encode the absolute timestamp of each \oursmodel{} token.
These tokens are concatenated and fed into the LLM.
To use the limited audio token budget more effectively, we propose \textit{transition-aware dynamic sampling} based on the transition probabilities.
Instead of uniformly sampling tokens over the full track, the model allocates more tokens to regions with high transition probability while preserving global coverage.
The model is then trained to generate timestamped captions, learning both timestamp localization and segment-level music description.
More details are described in \S\ref{app:caption_pretraining}.

\tightparagraph{Stage 3: QA Fine-Tuning.}
Next, we fine-tune the model with LoRA on the five QA tasks on \ours{} training set.
This stage adapts the model from timestamped captioning to five temporal understanding tasks.
Because the QA tasks have heterogeneous answer formats, we use an answer-only, sample-normalized, task-balanced objective so that long-form description tasks do not dominate training.
The full objective is provided in \S\ref{app:sft_objective}.

\tightparagraph{Stage 4: GRPO Training.}
Finally, we apply Group Relative Policy Optimization (GRPO)~\citep{shao2024deepseekmath} to timestamp-centric tasks.
Unlike token-level cross-entropy, GRPO allows us to optimize task-level rewards that directly reflect temporal grounding quality.

For TSG, which requires a single timestamp prediction, we combine the binary Hit@3s metric with a continuous distance-based reward:
\[
\begin{aligned}
r_{\mathrm{TSG}}
={}&
0.8 \cdot \mathbf{1}[d \leq 3]
+ 0.2 \cdot \exp\left(-\frac{d}{15}\right) \\
&- 0.5 \cdot \mathbf{1}_{\mathrm{out}}
- 1.0 \cdot \mathbf{1}_{\mathrm{fmt}} .
\end{aligned}
\]
where \(d = |\hat{t} - t^\ast|\) denotes the absolute temporal error between the predicted and gold timestamps.
The binary term directly optimizes the Hit@3s evaluation criterion and constitutes the primary reward signal.
We additionally introduce the continuous term to provide graded feedback outside the 3-second tolerance window by assigning higher rewards to predictions closer to the gold timestamp.
This alleviates reward sparsity caused by assigning the same binary reward to all predictions beyond the threshold and provides a smoother learning signal.
$\mathbf{1}_{\mathrm{out}}$ and $\mathbf{1}_{\mathrm{fmt}}$ penalize out-of-range and invalid-format predictions.

For MTR, which requires temporal interval prediction, we use a strict reward combining temporal IoU and F1.
We directly compare the predicted temporal intervals $\mathcal{P}$ with the gold intervals $\mathcal{G}$.
The reward is defined as:
\[
\begin{aligned}
r_{\mathrm{MTR}}
&=
0.5 \cdot \mathrm{IoU}(\mathcal{P},\mathcal{G})
+
0.5 \cdot \mathrm{F1}(\mathcal{P},\mathcal{G}) \\
&\quad
-
0.5 \cdot \mathbf{1}_{\mathrm{out}}
-
1.0 \cdot \mathbf{1}_{\mathrm{fmt}},
\end{aligned}
\]
Out-of-range predictions and invalid formats follow the same penalty scheme as TSG. This strict reward encourages the model to accurately localize the relevant temporal regions while producing precise interval predictions.

\section{Experiments}

\newcommand{\cmark}{\textcolor{green!70!black}{\ding{51}}}
\newcommand{\xmark}{\textcolor{red!80!black}{\ding{55}}}

\begin{table*}[t]
\centering
\small
\setlength{\tabcolsep}{3pt}
\resizebox{\textwidth}{!}{%
\begin{tabular}{llccc|cc|c|cc|c|cc|c}
\toprule
\multicolumn{5}{c|}{\textbf{Configuration}}
& \multicolumn{2}{c|}{\textbf{TSG}}
& \textbf{LTR}
& \multicolumn{2}{c|}{\textbf{TAD}}
& \textbf{GTO}
& \multicolumn{2}{c|}{\textbf{MTR}}
& \textbf{Total} \\
\cmidrule(lr){1-5}
\cmidrule(lr){6-7}
\cmidrule(lr){8-8}
\cmidrule(lr){9-10}
\cmidrule(lr){11-11}
\cmidrule(lr){12-13}
\cmidrule(lr){14-14}
\textbf{Ablation}
& \textbf{Setting}
& \makecell{\textbf{Caption}}
& \makecell{\textbf{QA}}
& \textbf{GRPO}
& \makecell{\textbf{Onset} \\ \textbf{Hit@3s}}
& \makecell{\textbf{Offset} \\ \textbf{Hit@3s}}
& \textbf{Acc.}
& \textbf{METEOR}
& \makecell{\textbf{CLAP} \\ \textbf{Score}}
& \textbf{Acc.}
& \makecell{\textbf{Temporal} \\ \textbf{IoU}}
& \makecell{\textbf{Temporal} \\ \textbf{F1}}
& \textbf{Avg.} \\
\midrule

\multirow{5}{*}{\makecell[l]{Training \\ stage}}
& Base model
& \xmark & \xmark & \xmark
& 39.0 & 3.5 & 45.7 & 9.2 & 33.7 & 46.5 & 6.4 & 10.6 & 24.3 \\

& Caption only
& \cmark & \xmark & \xmark
& 28.5 & 4.0 & 38.9 & 15.7 & 32.8 & 28.8 & 9.3 & 11.7 & 21.2 \\

& QA only
& \xmark & \cmark & \xmark
& 13.5 & 0.0 & 62.0 & 17.5 & 27.7 & 51.0 & 10.9 & 13.4 & 24.5 \\

& Caption + QA$^\dagger$
& \cmark & \cmark & \xmark
& 61.5 & 64.5 & 74.0 & 23.1 & 30.3 & 61.1 & 15.5 & 20.5 & 43.8 \\

& Full
& \cmark & \cmark & \cmark
& 65.5 & 65.0 & 74.0 & 22.4 & 30.3 & 61.1 & 17.5 & 25.6 & 45.2 \\

\midrule

\multirow{3}{*}{\makecell[l]{\oursmodel{} \\ token rate}}
& 6.66 tokens/sec
& \cmark & \cmark & \xmark
& 61.5 & 62.0 & 77.4 & 23.0 & 29.7 & 60.1 & 14.0 & 19.3 & 43.4 \\

& 3.33 tokens/sec$^\dagger$
& \cmark & \cmark & \xmark
& 61.5 & 64.5 & 74.0 & 23.1 & 30.3 & 61.1 & 15.5 & 20.5 & 43.8 \\

& 0 tokens/sec
& \cmark & \cmark & \xmark
& 41.5 & 48.5 & 63.0 & 21.7 & 30.0 & 52.0 & 10.8 & 14.5 & 35.3 \\

\midrule

\multirow{2}{*}{\makecell[l]{Sampling \\ strategy}}
& Uniform sampling
& \cmark & \cmark & \xmark
& 48.5 & 46.0 & 69.2 & 21.6 & 30.5 & 54.0 & 12.5 & 16.5 & 37.4 \\

& Dynamic sampling$^\dagger$
& \cmark & \cmark & \xmark
& 61.5 & 64.5 & 74.0 & 23.1 & 30.3 & 61.1 & 15.5 & 20.5 & 43.8 \\

\bottomrule
\end{tabular}%
}
\vspace{-0.5em}
\caption{\red{
Ablation on training stages, \oursmodel{} token rate, and sampling strategy.
The 0-token setting removes the \oursmodel{} tokens and relies only on the Qwen2.5-Omni audio encoder. $^\dagger$ inicates same model.}
}
\vspace{-1em}
\label{tab:ablation_study}
\end{table*}


In addition to the baseline analysis in \S\ref{sec:baseline_performance}, we evaluate our models on \ours{}.
Implementation details are provided in \S\ref{app:experimental_details}.
\red{As shown in Table~\ref{tab:main_results}, \oursmodel{} 7B achieves the highest overall score of 45.2, outperforming its corresponding base model by 20.9 percentage points (pp).
The largest improvement is observed in offset localization, with a gain of 61.5 pp, addressing one of the most challenging aspects of temporal grounding for baseline models.
Notably, \oursmodel{} 3B achieves an even larger overall improvement of 25.5 pp over its base model.
Furthermore, applying our training pipeline to the Phi-4-mm backbone yields a 19.0 pp improvement in overall performance.
These results show the effectiveness of our training pipeline across different model architectures and scales.}

\subsection{Ablation Studies}

\paragraph{Multi-Stage Training.}
We analyze the contribution of each training stage using Qwen2.5-Omni-7B as the backbone, including timestamped caption pretraining, QA fine-tuning, and GRPO training.

As shown in Table~\ref{tab:ablation_study}, caption pretraining alone decreases the overall score by 3.1 pp, as the model has not yet been adapted to the QA format.
QA fine-tuning alone improves performance on the multiple-choice tasks (LTR and GTO), but substantially degrades TSG performance, resulting in only a marginal improvement of 0.2 pp.
In contrast, combining caption pretraining with QA fine-tuning substantially boosts the overall score to 43.8, highlighting their complementary roles in learning temporal alignment and downstream QA capabilities.
Finally, adding GRPO achieves the best overall score of 45.2.
Beyond this aggregate improvement, GRPO also improves the validity of timestamp predictions.
Before GRPO, the model sometimes generates timestamps outside the valid input audio duration.
As shown in Figure~\ref{fig:grpo_out_of_range}, these out-of-range predictions decrease as GRPO training progresses.
This suggests that timestamp-aware preference optimization calibrates \oursmodel{} toward temporally valid outputs, encouraging predictions that are not only close to the target timestamps or intervals but also constrained by the actual audio duration.
Overall, these results show that the training stages play complementary roles, where caption pretraining supports temporal alignment, QA fine-tuning provides task-specific reasoning ability, and GRPO improves the temporal validity of timestamp-centric predictions.

\begin{figure}[t]
\centering
  \includegraphics[width=0.9\columnwidth]{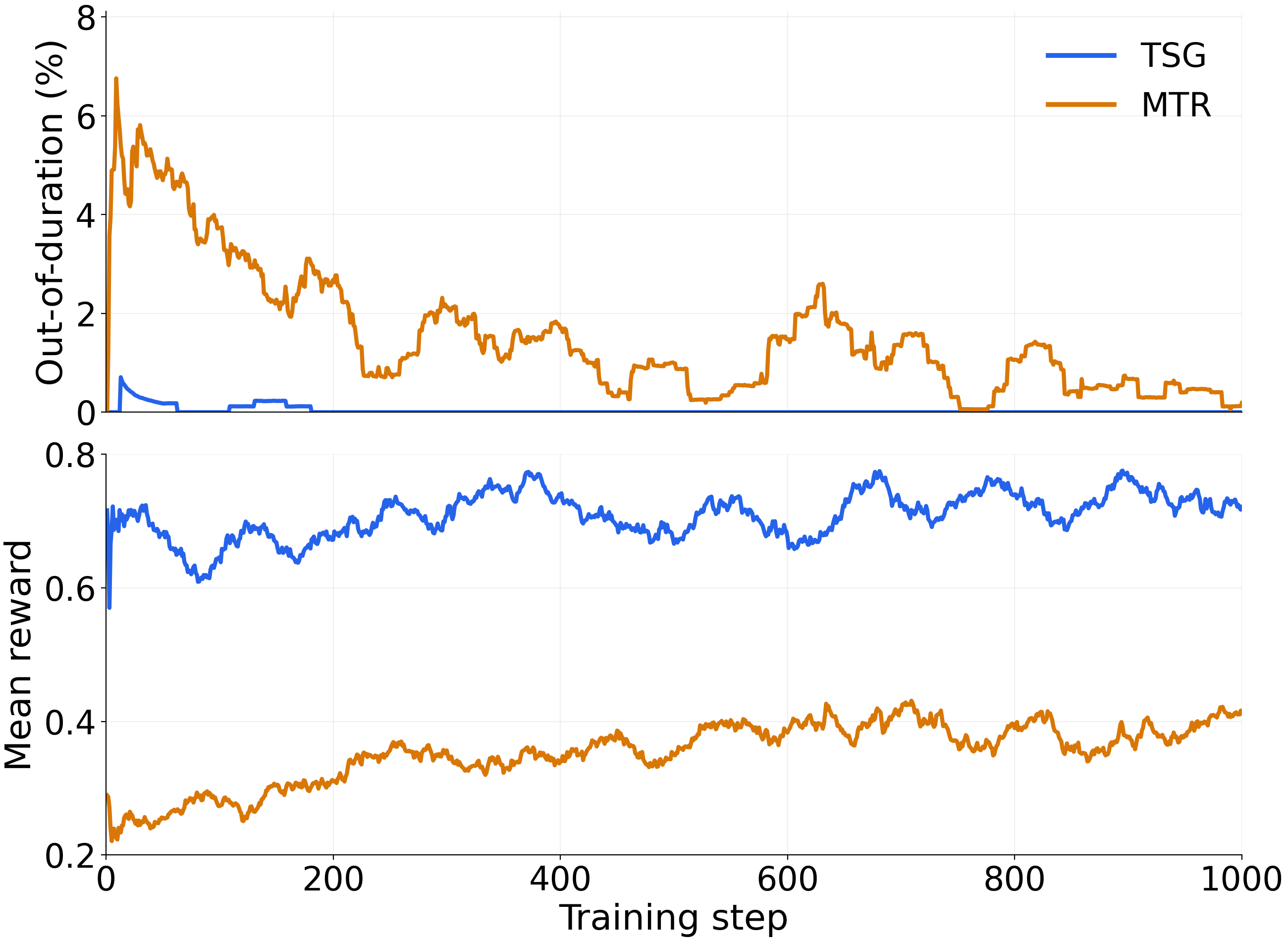}
  \caption{\red{
(Top) Percentage of generated rollouts containing timestamp predictions outside the audio duration. (Bottom) Mean reward during GRPO training.}
}
  \vspace{-15pt}
  \label{fig:grpo_out_of_range}
\end{figure}

\tightparagraph{\oursmodel{} Tokens.}
\red{
To examine the contribution of the \oursmodel{} tokens in our dual-encoder architecture, we vary the token rate of the \oursmodel{} encoder during caption pretraining and QA fine-tuning.
We compare 3.33 and 6.66 tokens/sec, corresponding to maximum budgets of 1000 and 2000 tokens for a 5 minute audio.
As shown in Table~\ref{tab:ablation_study}, adding \oursmodel{} tokens yields about 8 pp improvement over the setting without \oursmodel{} tokens. The 3.33 and 6.66 tokens/sec settings achieve comparable performance, with no benefit from further increasing the token rate, suggesting that 3.33 tokens/sec already provides sufficient temporal resolution for effective temporal grounding. Given that the original Qwen2.5-Omni audio encoder uses 25 tokens/sec, these results demonstrate that our \oursmodel{} tokens provide substantial temporal grounding gains with only a small number of additional tokens.}

\tightparagraph{Dynamic Sampling.}
\red{
To evaluate the contribution of our transition probability-guided dynamic sampling, we replace it with uniform sampling while keeping all other settings unchanged.
As shown in Table~\ref{tab:ablation_study}, dynamic sampling improves the overall score from 37.4 to 43.8.
The gains are particularly clear on tasks that require fine-grained access to musically important regions (TSG and LTR), where allocating more \oursmodel{} tokens around transition points is likely to be beneficial.
These results suggest that transition-aware dynamic sampling better captures temporally localized musical changes than uniform sampling.}

    
    


\section{Conclusion}

We introduced \ours{}, a benchmark for evaluating temporal grounding of LALMs.
Through five temporal grounding tasks, \ours{} reveals that existing LALMs still lack reliable temporal grounding ability in music understanding.
\red{To address this limitation, we further proposed \oursmodel{}, a novel four-stage temporal optimization recipe.
Experiments demonstrate that \oursmodel{} consistently improves temporal grounding over the baselines.
Overall, \ours{} highlights temporal grounding as a key challenge for future research and provides a benchmark and training recipe that may support the development of models that better capture not only what happens in music, but also when it happens.}

\section*{Limitations}\label{sec:limitations}

This work has four limitations.
First, \ours{} relies on an automatic data generation pipeline, so residual noise may remain despite multiple validation steps. Structural boundaries may miss musically meaningful transitions, and automatic components may introduce subtle errors. 
Second, mood trajectory reasoning is subjective, as perceived arousal can vary across listeners, genres, and contexts. Thus, the arousal intervals in \ours{} should be viewed as benchmark-specific annotations rather than definitive judgments of musical emotion.
\red{
Third, while the \ours{} benchmark and all associated resources, including metadata and the full data-generation pipeline and code, are publicly released, the raw audio files from an internal dataset used for training data generation cannot be redistributed. Nevertheless, the released resources allow researchers to reproduce the data construction process using their own permissible or custom audio collections.
\hypertarget{paper:limitation-coupling-lines}{}%
Finally, although derived from distinct source audios, since \ours{} and the \oursmodel{} training data share task definitions and related construction procedures, the observed gains may partly reflect alignment with the benchmark’s task formats and annotation style. Our results demonstrate improvements under the \ours{} evaluation setting, while generalization to independently constructed music temporal grounding tasks remains to be established.}

\section*{Ethical Considerations}

\ours{} is intended for research on temporally grounded music understanding. Since it is derived from existing music datasets and annotations, all released annotations, metadata, evaluation scripts, and processing code should comply with the licenses and access conditions of the original datasets, models, and tools; restricted audio content will not be redistributed. The benchmark focuses on musical events and affective changes rather than personal identity, but recordings with vocals should not be used to identify singers or infer sensitive attributes. Model outputs may contain incorrect timestamps or descriptions and should not be used as authoritative evidence in copyright, forensic, or commercial rights settings without expert verification. \ours{} may also inherit dataset and tool biases across genres, instruments, languages, and cultural traditions. AI-based writing assistance was used only for language polishing and clarity, while all research ideas, design decisions, experiments, analyses, and conclusions were developed and verified by the authors.

\bibliography{custom}

\clearpage

\appendix

\section{Appendix}

\label{sec:appendix}

\subsection{Overview}

This appendix provides additional details on benchmark construction, model training, evaluation protocols, and supplementary analyses.

\begin{itemize}[leftmargin=*, noitemsep, topsep=0pt]
    \item \textbf{Timestamped music caption construction. (\S\ref{sec:appendix_timestamped_caption_generation})}
    We describe the construction pipeline for source audio filtering and transition clip extraction, mood-change annotation, feature-grounded captioning, caption cross-validation.

    \item \textbf{Human-assisted Validation and Annotator Details. (\S\ref{sec:appendix_human_validation})}
    We provide details on the human-assisted validation process used to finalize the QA dataset, as well as information about the human annotators who participated in this work.

    \item \textbf{Model training. (\S\ref{sec:appendix_model_training_details})}
    We provide implementation details for transition-aware encoder pretraining, timestamped caption pretraining, and supervised QA fine-tuning.

    \item \textbf{Additional analyses and experimental details. (\S\ref{sec:appendix_additional_ablations})}
    We include supplementary analyses such as qualitative examples, dataset statistics, feature statistics, baseline inference protocols, and training hyperparameters.
\end{itemize}

\subsection{Timestamped Caption Generation Details}\label{sec:appendix_timestamped_caption_generation}

This section provides additional details on the construction of timestamped music caption generation.

\subsubsection{Source Audio Filtering and Transition Clip Extraction}
\label{sec:appendix_2.1}

\paragraph{Source Audio Filtering.}
We use an internal music dataset as the source audio collection, which contains 21K full-track music recordings.
From this collection, we retain 17K tracks with durations between 1 and 5 minutes.
This duration range is chosen because the tracks are long enough to contain multiple musical events and structural changes, while remaining computationally tractable for dense segmentation, caption generation, and downstream QA construction.

\paragraph{Structural Segmentation and Transition Clip Extraction.}
To obtain candidate timestamps for musically meaningful changes, we apply a structural music segmentation model~\citep{hao2025songformer} to each track.
The resulting segment boundaries are treated as transition timestamps.
For each boundary at timestamp $T$, we extract a 20-second transition clip spanning $[T-10, T+10]$, where the transition point is centered in the clip.
These transition clips are used for mood-change annotation, mood-change prediction, and dynamic caption generation.


\begin{table}[t]
\centering
\small
\resizebox{\columnwidth}{!}{%
\begin{tabular}{@{} l c c c c c c @{}}
\toprule
&
\multicolumn{1}{c}{\textbf{TSG}}
&
\multicolumn{4}{c}{\textbf{Other Temporal QA Tasks}}
&
\textbf{Total} \\
\cmidrule(r){2-2}
\cmidrule(lr){3-6}

& \textbf{TSG}
& \textbf{LTR}
& \textbf{TAD}
& \textbf{GTO}
& \textbf{MTR}
& \\

\midrule
\multicolumn{7}{@{}l}{\textit{Data Source}} \\[2pt]

Train
& S+I
& I & I & I & I
& -- \\

Test
& S+M
& M & M & M & M
& -- \\

\midrule
\multicolumn{7}{@{}l}{\textit{Number of Examples}} \\[2pt]

Train
& 8,000
& 8,000 & 8,000 & 8,000 & 8,000
& \textbf{40,000} \\

Test
& 400
& 208 & 208 & 198 & 250
& \textbf{1,264} \\

\bottomrule
\end{tabular}%
}
\caption{\red{
Task-wise dataset sources and split statistics.
S, I, and M denote Slakh2100, internal dataset, and MTG-Jamendo, respectively.}
}
\label{tab:appendix_train_statistics}
\end{table}

\begin{table}[t]
\centering
\small
\resizebox{\columnwidth}{!}{%
\begin{tabular}{lcccc}
\toprule
\textbf{Category}
& \makecell{\textbf{Tracks with} \\ $\mathbf{\geq 1}$ \textbf{tag}}
& \textbf{Unique Tags}
& \textbf{Tag Counts}
& \makecell{\textbf{Avg. Tags} \\ \textbf{/ Track}} \\
\midrule
Genre
& 745 & 154 & 1,902 & 2.55 \\

Instrument
& 342 & 76 & 991 & 2.90 \\

Mood/Theme
& 299 & 137 & 610 & 2.04 \\
\midrule
Total
& 749 & 367 & 3,503 & 4.68 \\
\bottomrule
\end{tabular}
}
\caption{\red{
Statistics of metadata tags by category, including genre, instrument, and mood/theme annotations.}
}
\label{tab:benchmark_diversity_analysis}
\end{table}

\begin{table*}[t]
\centering
\small
\resizebox{0.7\textwidth}{!}{%
\begin{tabular}{ll}
\toprule
\textbf{Category} & \textbf{Top 5 Tags (Track Count)} \\
\midrule
Genre
& electronic (201), pop (129), soundtrack (108), rock (107), ambient (85) \\

Instrument
& piano (121), synthesizer (97), bass (83), drums (81), guitar (63) \\

Mood/Theme
& emotional (24), happy (23), film (22), love (21), relaxing (21) \\
\bottomrule
\end{tabular}%
}
\caption{\red{
Top five most frequent metadata tags in each category, reported by track count.}
}
\label{tab:benchmark_tag_analysis}
\end{table*}

\subsubsection{Mood-Change Annotation Details}
\label{sec:appendix_mood_annotation}

\red{We use the mood prediction model provided by Essentia \texttt{deam-msd-musicnn}~\citep{bogdanov2013essentia}, trained on the DEAM dataset~\citep{aljanaki2017developing}, to estimate musical arousal on an absolute scale from 1 to 9. For each transition, we compute the average arousal within windows before and after the transition timestamp and define the arousal difference as the post-transition average minus the pre-transition average. Negative values therefore indicate a decrease in musical intensity after the transition, whereas positive values indicate an increase.}

\subsubsection{Feature-Grounded Captioning Details}
\label{sec:appendix_feature_captioning}

\paragraph{Feature Extraction.}
For each structural segment, we extract musical features that can provide grounded evidence for caption generation.
These include tempo, key, and volume-related features computed using standard audio analysis libraries, including librosa~\citep{mcfee2015librosa}, madmom~\citep{bock2016madmom}, and Essentia~\citep{bogdanov2013essentia}.
These features are used to constrain the captioning process so that generated descriptions reflect measurable properties of the audio rather than relying only on free-form model priors.
The extracted feature set is summarized in Table~\ref{tab:extracted_features}.

In addition to global musical features, we apply a source separation model~\citep{rouard2023hybrid} to obtain stem-level signals.
The separation model provides coarse 4-stem outputs (vocals, drums, bass, and others).
These source-level cues provide evidence about the presence and activity of major musical sources within each segment, which helps reduce hallucinations in instrument- and vocal-related descriptions.

\paragraph{Static Caption Generation.}
For each segment, we provide the extracted musical features and source-level cues as grounded inputs to the caption generation model \red{(Qwen3-Omni-30B-A3B)~\cite{xu2025qwen3}.}
The model generates a segment-level static caption that describes the musical state within the segment, including instrumentation, rhythm, harmony, texture, and mood when supported by the extracted features.
These static captions are aligned with segment-level timestamp intervals and are intended to describe what is present within each segment.

\paragraph{Dynamic Caption Generation.}
For each segment boundary, we compare the features extracted from the pre-boundary and post-boundary segments.
We also include the predicted mood-change score from the mood-change predictor.
Using these before/after features, the caption generation model produces a dynamic caption describing how the music changes across the transition.
These dynamic captions capture localized musical changes such as instrument entries or exits, rhythmic intensification, texture reduction, harmonic shifts, and changes in emotional intensity.

\paragraph{Interleaving Static and Dynamic Captions.}
We combine static and dynamic captions into a single timestamped caption sequence.
Static captions are associated with segment intervals, while dynamic captions are associated with transition timestamps.
This produces an interleaved description of the full track, where relatively stable musical states and local transition events are both explicitly grounded in time.

\paragraph{Sliding-Window Rewriting.}
Because static and dynamic captions are initially generated independently, the resulting caption sequence can lack coherence when read as a full-track description.
To address this issue, we apply a sliding-window rewriting step using gpt-oss-120b~\citep{agarwal2025gpt}.
The rewriting model sequentially refines neighboring captions so that the final timestamped caption reads as a coherent description of the full track.
During rewriting, the model is instructed to preserve the original musical facts, timestamps, and temporal structure, while improving fluency and local coherence.
Examples before and after rewriting are shown in Figure~\ref{fig:timestamped_caption_before_rewriting} and Figure~\ref{fig:timestamped_caption_after_rewriting}.

\subsubsection{Caption Cross-Validation Details}
\label{sec:appendix_cross_validation}
An initial human validation study revealed that some generated captions contained errors in instrument and key descriptions.
For instrument descriptions, the source separation model provides only coarse 4-stem outputs, which are insufficient for distinguishing fine-grained instruments such as guitar or piano.
\red{We therefore train a MERT-based classifier on the 20 instrument classes of OpenMIC-2018~\citep{humphrey2018openmic}, then we group these labels into nine broader instrument categories.}
The classifier is used to verify instrument mentions in generated captions, and unsupported descriptions are revised before QA generation.
For key descriptions, we apply a confidence-based verification rule.
A key mention is retained when \red{the classifier~\cite{bock2016madmom}} assigns a probability of at least 0.9 to the mentioned key, and removed when the probability is below 0.1.
For intermediate confidence scores, we keep the original caption unchanged.
Tables~\ref{tab:instrument_distribution_before} and~\ref{tab:instrument_distribution_after} show statistics of added/removed instruments during cross-validation.

\subsection{Human-assisted Validation and Annotator Details}
\label{sec:appendix_human_validation}

\paragraph{Human-assisted Validation}
We apply task-specific validation procedures to ensure the reliability of the final benchmark annotations. Detailed instructions are included in Figures~\ref{fig:human_annotation_instruction_type2}, 
\ref{fig:human_annotation_instruction_type4}, 
and~\ref{fig:human_annotation_instruction_type5}.

For TSG, we skip additional human validation because the ground-truth timestamps are obtained from MIDI-aligned instrument tracks and separated vocal tracks. 
We further retain only examples where the target source is clearly audible according to a volume-based filtering criterion. 
Since the resulting onset and offset timestamps are derived from structured source-level annotations and filtered for perceptual salience, additional human correction is not required.

For LTR and TAD, we provide annotators with the full audio, the transition timestamp $T$, and the corresponding transition description. 
Annotators are asked to verify whether the description accurately matches the musical change occurring around $T$. 
If the description contains incorrect or unsupported musical content, they are instructed to revise it so that it correctly reflects the audio at the given timestamp.

For GTO, we provide annotators with three transition descriptions $(X, Y, Z)$ along with their corresponding timestamps. 
Annotators first verify whether each description is temporally aligned with the actual musical change in the audio. 
If any description is inaccurate or misaligned, they are instructed to correct it before the chronological ordering question is finalized. 
This ensures that the task evaluates temporal ordering rather than errors in the transition descriptions themselves.

For MTR, music experts directly listen to the audio and annotate the temporal regions with the highest or lowest arousal, depending on the question. \red{The inter-annotator correlation for arousal reached 0.8, demonstrating substantial consensus among the experts. To account for the inherent subjectivity of emotional perception}, annotators were allowed to provide up to four valid duration intervals when multiple regions exhibited comparable arousal levels. These expert annotations are used as the ground-truth intervals for evaluating mood trajectory reasoning.

\begin{table}[t]
\centering
\small
\resizebox{\columnwidth}{!}{%
\begin{tabular}{lcccc}
\toprule
\textbf{Task}
& \textbf{Candidates}
& \textbf{Accepted}
& \makecell{\textbf{Minor} \\ \textbf{Revision}}
& \textbf{Rejected} \\
\midrule
LTR / TAD
& 255
& 137 (53.7\%)
& 71 (27.8\%)
& 47 (18.4\%) \\

GTO
& 198
& 16 (8.1\%)
& 182 (91.9\%)
& 0 (0.0\%) \\
\bottomrule
\end{tabular}
}
\caption{\red{
Number and proportion of candidate instances accepted, revised, or rejected by expert annotators.}
}
\label{tab:expert_validation}
\end{table}

\red{In Table~\ref{tab:expert_validation}, we report the number and proportion of candidate instances accepted, revised, or rejected by expert annotators to better quantify annotation noise. For GTO, the low acceptance rate is due to our stricter criterion that all three transition descriptions must be correct. Importantly, there were no cases where the temporal ordering itself was incorrect. Therefore, no GTO candidates were rejected for incorrect ordering.}

\paragraph{Human Annotation.}
Human annotation and validation were conducted by temporary research support staff affiliated with our institution. 
The annotators were not recruited through a public crowdsourcing platform, and their duties were not limited to this particular annotation assignment. 
Their employment and compensation complied with local labor laws and institutional procedures. 
Annotators were informed that their annotations would be used for research purposes. 
No personal identifying information about annotators is included in the released artifacts or reported analyses.

All members have graduated from a music university or a specialized music school.
In terms of professional experience, each individual has experience in at least one of the following: performance, composition, arrangement, serving as a part-time lecturer at a music university, conducting research at a university, or audio analysis.

\subsection{Model Training Details}\label{sec:appendix_model_training_details}

\subsubsection{Transition-Aware Encoder Pretraining Details}
\label{app:encoder_pretraining}

We initialize the transition-aware music encoder from \red{MERT-v1-330M~\citep{li2024mert}} and adapt it with LoRA~\citep{hu2021lora}.
The encoder is trained with two auxiliary temporal objectives: transition probability prediction and mood-change prediction.

For transition probability prediction, we use structural segment boundaries obtained from the music segmentation model as supervision.
Rather than treating each boundary as a hard frame label, we construct a Gaussian target distribution centered at each boundary timestamp.
A lightweight 1D convolutional prediction head is attached to the frame-level MERT representations to estimate transition probabilities over time.
This objective encourages the encoder to assign high probability to musically salient change points, such as source entries, rhythmic changes, and structural transitions.

For mood-change prediction, we attach a linear regression head to the encoder and train it to predict the change in emotional intensity around each transition boundary.
The supervision comes from the mood prediction model as described in \S\ref{sec:appendix_mood_annotation}.
This objective encourages the encoder to capture affective temporal dynamics in addition to structural changes.

\subsubsection{Timestamped Caption Pretraining Details}
\label{app:caption_pretraining}

In timestamped caption pretraining, the model uses two complementary audio streams.
The frozen backbone audio encoder provides semantic audio representations, while the transition-aware \oursmodel{} encoder provides fine-grained acoustic and temporal representations.
Because the two encoders produce representations in different embedding spaces, we use a learnable projector to map \oursmodel{} encoder outputs into the LLM embedding space.

We add sinusoidal time embeddings to the projected \oursmodel{} tokens based on their absolute timestamps in the original audio.
This allows the LLM to distinguish not only the content of each token but also when it occurs in the track.
The projected \oursmodel{} tokens and backbone audio tokens are concatenated sequentially, separated by a special \texttt{<AUDIO\_SPLIT>} token, and then provided to the LLM together with the text prompt.

As shown in Figure~\ref{fig:dynamic_pooling}, to fit long music tracks into a limited token budget, we apply transition-aware dynamic sampling using the transition probabilities predicted by the encoder.
Given $T$ encoder frames at a frame rate of $f=75$ frames per second, with transition probabilities $\{p_t\}_{t=1}^{T}$, we select $K \leq T$ frames, where $K$ is determined by the track duration and the maximum token budget.
The transition probabilities control the temporal allocation of these tokens, assigning more tokens to likely transition regions while preserving coverage over the full track.

We first smooth the transition probabilities with a Gaussian kernel $G_\sigma$ and apply a power transformation:
\begin{equation}
\begin{aligned}
    \tilde{p}_t &= (G_\sigma * p)_t, \\
    q_t &=
    \frac{(\tilde{p}_t + \epsilon)^\gamma}
         {\sum_{j=1}^{T}(\tilde{p}_j + \epsilon)^\gamma},
\end{aligned}
\end{equation}
where $\epsilon$ ensures numerical stability and $0 < \gamma < 1$ reduces excessive concentration around sharp probability peaks.
We use a Gaussian standard deviation of $\sigma=2.0$ frames (approximately $26.7$~ms) and set $\gamma=0.7$.
We then mix the transition-based distribution with a uniform distribution:
\begin{equation}
    w_t = (1-\alpha)q_t + \frac{\alpha}{T},
    \label{eq:transition_sampling_weights}
\end{equation}
where $\alpha \in [0,1]$ controls the contribution of uniform coverage.
We set $\alpha=0.2$, assigning mixture weights of $80\%$ to the transition-based distribution and $20\%$ to the uniform distribution.
To further moderate local concentration, we apply a neighborhood-based penalty and renormalize:
\begin{equation}
\begin{aligned}
    \ell_t &= \operatorname{AvgPool}_{W}(w)_t, \\
    \hat{w}_t
    &=
    \frac{
        w_t/(1+\lambda\ell_t)
    }{
        \sum_{j=1}^{T} w_j/(1+\lambda\ell_j)
    },
\end{aligned}
\end{equation}
where $W$ is the averaging window size and $\lambda$ controls the penalty strength.
We use $W=60$ frames, corresponding to $0.8$~s, and set $\lambda=3.0$.

We select frames using evenly spaced quantiles of the cumulative distribution:
\begin{equation}
\begin{aligned}
    F(t) &= \sum_{j=1}^{t}\hat{w}_j, \\
    t_k
    &=
    \min\left\{
        t :
        F(t) \geq
        \frac{k+\tfrac{1}{2}}{K}
    \right\}, \\
    &\hspace{2.5cm} k=0,\ldots,K-1.
\end{aligned}
\label{eq:transition_inverse_cdf}
\end{equation}
This deterministic inverse-CDF procedure yields a higher sampling density in regions with larger transition weights.
At the encoder frame rate $f$, the corresponding local token rate is approximately
\begin{equation}
    r_t \approx K f \hat{w}_t.
    \label{eq:transition_token_rate}
\end{equation}
Thus, the total token budget remains fixed for a given track, while the local temporal resolution adapts to the predicted transitions.
Finally, we remove duplicate indices and prioritize temporally separated frames, filling any remaining budget with the highest-weight unselected frames.
The selected tokens retain their original timestamps for the sinusoidal time embeddings.

\begin{figure}[t]
  \includegraphics[width=\columnwidth]{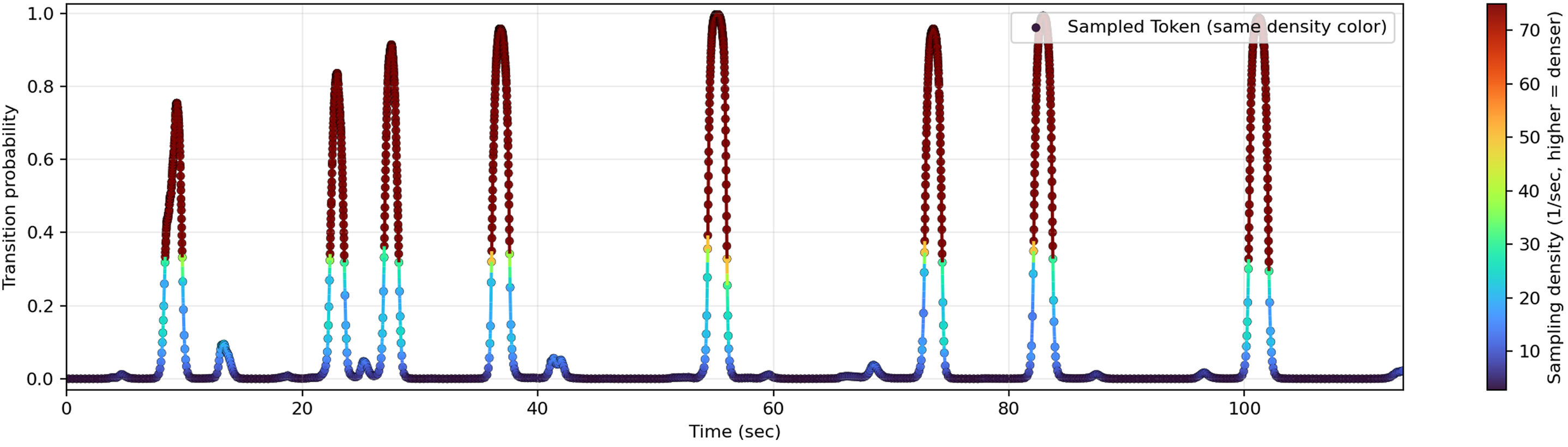}
  \caption{\red{Dynamic sampling strategy focusing more tokens on areas with high transition probability.}}
  \label{fig:dynamic_pooling}
\end{figure}

\subsubsection{Supervised QA Fine-Tuning Objective}
\label{app:sft_objective}

Our QA fine-tuning data contain heterogeneous answer formats, including multiple-choice letters, free-form descriptions, single timestamps, and temporal intervals.
Because these answer types have substantially different sequence lengths, a standard token-averaged language modeling loss can overemphasize tasks with longer answers, such as transition-aware description.
We therefore use an answer-only, sample-normalized, task-balanced supervised fine-tuning objective.

For each training example $i$, we mask all audio tokens, prompt tokens, instruction tokens, and question tokens with the ignore label, and compute the loss only on the answer tokens.
Let $\mathcal{A}_i$ denote the set of supervised answer-token positions for example $i$, and let $\ell_{i,j}$ be the token-level negative log-likelihood at answer position $j$.
We first normalize the loss within each sample:
\[
\mathcal{L}_i
=
\frac{1}{|\mathcal{A}_i|}
\sum_{j \in \mathcal{A}_i}
\ell_{i,j}.
\]

To prevent tasks with more examples in a minibatch from dominating the update, we then average losses within each QA type and assign equal weight to all QA types present in the minibatch.
Let $\mathcal{B}_k$ denote the set of examples of QA type $k$ in the minibatch, and let $\mathcal{T}_{\mathcal{B}}$ denote the set of QA types that appear in the minibatch.
The final supervised fine-tuning loss is:
\[
\begin{aligned}
\mathcal{L}_{\mathrm{SFT}}
&=
\frac{1}{|\mathcal{T}_{\mathcal{B}}|}
\sum_{k \in \mathcal{T}_{\mathcal{B}}}
\frac{1}{|\mathcal{B}_k|}
\sum_{i \in \mathcal{B}_k}
\mathcal{L}_i \\
&=
\frac{1}{|\mathcal{T}_{\mathcal{B}}|}
\sum_{k \in \mathcal{T}_{\mathcal{B}}}
\frac{1}{|\mathcal{B}_k|}
\sum_{i \in \mathcal{B}_k}
\frac{1}{|\mathcal{A}_i|}
\sum_{j \in \mathcal{A}_i}
\ell_{i,j}.
\end{aligned}
\]

This objective ensures that supervision is applied only to the target answer. Each example is normalized by answer length, and each QA type present in the minibatch receives equal total weight.

\subsection{Additional Analyses and Experimental Details}\label{sec:appendix_additional_ablations}

\subsubsection{Additional Analyses}
\label{app:additional_ablations}

\paragraph{Effect of Zero-shot CoT Prompting.}
\begin{table*}[t]
\centering
\small
\setlength{\tabcolsep}{5pt}
\resizebox{\textwidth}{!}{%
\begin{tabular}{l|cc|c|cc|c|cc|c}
\toprule
\textbf{Prompting Strategy}
& \multicolumn{2}{c|}{\textbf{TSG}}
& \textbf{LTR}
& \multicolumn{2}{c|}{\textbf{TAD}}
& \textbf{GTO}
& \multicolumn{2}{c|}{\textbf{MTR}}
& \textbf{Total} \\
\cmidrule(lr){2-3}
\cmidrule(lr){4-4}
\cmidrule(lr){5-6}
\cmidrule(lr){7-7}
\cmidrule(lr){8-9}
\cmidrule(lr){10-10}
& \makecell{\textbf{Onset} \\ \textbf{Hit@3s}}
& \makecell{\textbf{Offset} \\ \textbf{Hit@3s}}
& \textbf{Acc.}
& \textbf{METEOR}
& \textbf{CLAPScore}
& \textbf{Acc.}
& \makecell{\textbf{Temporal} \\ \textbf{IoU}}
& \makecell{\textbf{Temporal} \\ \textbf{F1}}
& \textbf{Avg.} \\
\midrule
Without CoT
& 61.5 & 64.5 & 74.0 & 23.1 & 30.3 & 61.1 & 15.5 & 20.5 & 43.8 \\

Zero-shot CoT
& 54.0 & 44.0 & 64.9 & 20.1 & 28.5 & 50.5 & 12.6 & 17.2 & 36.5 \\
\bottomrule
\end{tabular}%
}
\caption{\red{
Effect of zero-shot chain-of-thought prompting. 
We compare the default prompting strategy without CoT against zero-shot CoT prompting.}
}
\label{tab:cot_ablation}
\end{table*}
We additionally examine whether zero-shot chain-of-thought prompting~\citep{kojima2022large} improves temporal grounding performance.
Specifically, the model is instructed to output its thinking process within \texttt{<think>...</think>} tags before generating the final prediction.
As shown in Table~\ref{tab:cot_ablation}, zero-shot CoT decreases the overall score from 43.8 to 36.5, a drop of 7.3 pp.
Performance declines across all evaluation metrics, with the largest degradation observed in offset localization (20.5 pp).
These results suggest that zero-shot CoT prompting does not necessarily benefit temporal grounding and may hinder precise temporal predictions.

\paragraph{Effect of LoRA Rank.}



\begin{table*}[t]
\centering
\small
\setlength{\tabcolsep}{5pt}
\resizebox{\textwidth}{!}{%
\begin{tabular}{l|cc|c|cc|c|cc|c}
\toprule
\textbf{LoRA Rank}
& \multicolumn{2}{c|}{\textbf{TSG}}
& \textbf{LTR}
& \multicolumn{2}{c|}{\textbf{TAD}}
& \textbf{GTO}
& \multicolumn{2}{c|}{\textbf{MTR}}
& \textbf{Total} \\
\cmidrule(lr){2-3}
\cmidrule(lr){4-4}
\cmidrule(lr){5-6}
\cmidrule(lr){7-7}
\cmidrule(lr){8-9}
\cmidrule(lr){10-10}
& \makecell{\textbf{Onset} \\ \textbf{Hit@3s}}
& \makecell{\textbf{Offset} \\ \textbf{Hit@3s}}
& \textbf{Acc.}
& \textbf{METEOR}
& \textbf{CLAPScore}
& \textbf{Acc.}
& \makecell{\textbf{Temporal} \\ \textbf{IoU}}
& \makecell{\textbf{Temporal} \\ \textbf{F1}}
& \textbf{Avg.} \\
\midrule

LoRA 64
& 65.5 & 65.0 & 74.0 & 22.4 & 30.3 & 61.1 & 17.5 & 25.6 & 45.2 \\

LoRA 128
& 64.5 & 66.5 & 75.0 & 22.4 & 30.5 & 61.1 & 18.4 & 26.6 & 45.6 \\

LoRA 256
& 69.5 & 61.5 & 74.0 & 23.6 & 29.6 & 61.1 & 16.9 & 24.4 & 45.1 \\

\bottomrule
\end{tabular}%
}
\caption{\red{
Ablation study on the LoRA rank used during QA fine-tuning and GRPO training.}
}
\label{tab:lora_ablations}
\end{table*}
We further analyze the sensitivity of our model to the LoRA rank used during QA fine-tuning and GRPO training.
As shown in Table~\ref{tab:lora_ablations}, LoRA ranks of 64, 128, and 256 achieve comparable overall scores of 45.2, 45.6, and 45.1, respectively, with a maximum difference of only 0.5 pp.
These results suggest that our model is relatively insensitive to the LoRA rank.
We therefore adopt rank 64 as our main setting to reduce computational costs while maintaining competitive performance.

\tightparagraph{Music Downstream Tasks of \oursmodel{} Encoder.}
\noindent We evaluate whether \oursmodel{} encoder training preserves the general musical representations of the original MERT encoder.
For each encoder, we train only a task-specific linear head following~\citet{ghosh2025music}.
As shown in Table~\ref{tab:encoder_ablation_downstream}, the \oursmodel{} encoder achieves somewhat lower performance than MERT on both genre~\citep{tzanetakis2002musical} and key~\citep{knees2015two} classification tasks.
This decrease is expected, as our encoder is specifically optimized to capture localized musical changes around transition points, rather than static, track-level attributes such as genre or key.

\paragraph{Additional Evaluation of TAD.}
\begin{table}[t]
\centering
\small
\resizebox{\columnwidth}{!}{%
\begin{tabular}{clclc}
\toprule
\textbf{Rank}
& \textbf{Model (BERTScore)}
& \textbf{F1}
& \textbf{Model (LLM Judge)}
& \textbf{Score /10} \\
\midrule
1  & GPT Audio                 & 28.5 & Gemini 2.5 Pro           & 3.1 \\
2  & GPT Audio 1.5             & 28.1 & \textbf{\oursmodel{} Qwen2.5 7B} & \textbf{2.9} \\
3  & Gemini 2.5 Flash          & 27.2 & Gemini 2.5 Flash         & 2.8 \\
4  & \textbf{\oursmodel{} Qwen2.5 3B} & \textbf{25.4} & GPT Audio 1.5 & 2.7 \\
5  & \textbf{\oursmodel{} Qwen2.5 7B} & \textbf{25.3} & \textbf{\oursmodel{} Qwen2.5 3B} & \textbf{2.6} \\
6  & Qwen2.5 Omni 3B           & 25.1 & Music Flamingo 2601      & 2.3 \\
7  & Music Flamingo 2601       & 24.8 & GPT Audio                & 2.1 \\
8  & Gemini 2.5 Pro            & 24.8 & Gemini 3.1 Pro Preview   & 1.9 \\
9  & Gemini 3 Flash Preview    & 24.8 & Gemini 3 Flash Preview   & 1.9 \\
10 & Qwen2.5 Omni 7B           & 24.3 & Qwen2.5 Omni 7B          & 1.8 \\
11 & Gemini 3.1 Pro Preview    & 23.5 & Audio Flamingo Next      & 1.8 \\
12 & \textbf{\oursmodel{} Phi4} & \textbf{21.9} & Qwen2.5 Omni 3B & 1.8 \\
13 & Qwen3 Omni 30B A3B        & 21.5 & Phi-4 Multimodal         & 1.5 \\
14 & Audio Flamingo Next       & 20.0 & \textbf{\oursmodel{} Phi4} & \textbf{1.3} \\
15 & Phi-4 Multimodal          & 12.6 & Qwen3 Omni 30B A3B       & 1.0 \\
\bottomrule
\end{tabular}%
}
\caption{\red{
Additional evaluation results on the transition-aware description (TAD) task using BERTScore F1 and an LLM-based judge. BERTScore values are multiplied by 100, while LLM judge scores are reported on a 10-point scale. Models are independently ranked by each metric in descending order.}
}
\label{tab:additional_tad_evaluation}
\end{table}
To provide a more comprehensive evaluation of transition-aware description (TAD), we introduce two additional metrics beyond METEOR and CLAPScore. While METEOR primarily measures lexical overlap and CLAPScore evaluates audio-text similarity, neither directly captures the semantic correctness of generated descriptions. We therefore employ BERTScore~\cite{zhang2019bertscore} to measure semantic similarity with reference descriptions and an LLM-based judge using \texttt{gpt-oss-120b} to assess description quality.

As shown in Table~\ref{tab:additional_tad_evaluation}, our models demonstrate competitive performance under both metrics. In terms of BERTScore, \oursmodel{} 3B and 7B achieve scores of 25.4 and 25.3, ranking fourth and fifth among 15 models, respectively. Notably, \oursmodel{} 7B achieves the second-highest LLM judge score of 2.9, outperforming all models except Gemini 2.5 Pro. \oursmodel{} 3B also ranks fifth with a score of 2.6.

Interestingly, model rankings differ substantially between BERTScore and LLM-based evaluation, suggesting that semantic similarity to reference descriptions does not necessarily reflect the quality assessed by an LLM judge. These results highlight the importance of complementary evaluation metrics.

\paragraph{Statistical Significance.}
To assess the statistical reliability of our results, we compute 95\% confidence intervals using paired bootstrap resampling with 10,000 iterations.
For each comparison, we resample evaluation examples while preserving the pairing between the base model and \oursmodel{} predictions.
Table~\ref{tab:bootstrap_ci} reports the performance improvements and their corresponding confidence intervals.
Across all three backbones, the confidence intervals remain above zero for all metrics except CLAPScore, supporting statistically reliable improvements across the remaining evaluation metrics.

\begin{table*}[t]
\centering
\small
\setlength{\tabcolsep}{4pt}
\newcommand{\bootci}[3]{%
    \makecell{#1 \\ {\scriptsize [#2, #3]}}%
}
\resizebox{\textwidth}{!}{%
\begin{tabular}{lccccccccc}
\toprule
& \multicolumn{2}{c}{\textbf{TSG}}
& \textbf{LTR}
& \multicolumn{2}{c}{\textbf{TAD}}
& \textbf{GTO}
& \multicolumn{2}{c}{\textbf{MTR}}
& \textbf{Total} \\
\cmidrule(lr){2-3}
\cmidrule(lr){4-4}
\cmidrule(lr){5-6}
\cmidrule(lr){7-7}
\cmidrule(lr){8-9}
\cmidrule(lr){10-10}

\textbf{Model}
& \makecell{\textbf{Onset} \\ \textbf{Hit@3s}}
& \makecell{\textbf{Offset} \\ \textbf{Hit@3s}}
& \textbf{Acc.}
& \textbf{METEOR}
& \makecell{\textbf{CLAP} \\ \textbf{Score}}
& \textbf{Acc.}
& \makecell{\textbf{Temporal} \\ \textbf{IoU}}
& \makecell{\textbf{Temporal} \\ \textbf{F1}}
& \textbf{Avg.} \\
\midrule

\textbf{\oursmodel{} (Phi-4-mm)}
& \bootci{+6.0}{0.5}{12.0}
& \bootci{+34.0}{27.5}{40.5}
& \bootci{+38.4}{30.3}{46.2}
& \bootci{+10.3}{9.2}{11.4}
& \bootci{+1.4}{-1.5}{4.3}
& \bootci{+38.4}{30.3}{46.5}
& \bootci{+9.3}{7.7}{10.8}
& \bootci{+14.0}{11.8}{16.3}
& \bootci{\textbf{+19.0}}{17.1}{20.9} \\

\textbf{\oursmodel{} (Qwen 2.5 Omni 3B)}
& \bootci{+53.0}{46.0}{60.0}
& \bootci{+62.0}{55.0}{69.0}
& \bootci{+37.5}{28.8}{45.7}
& \bootci{+11.8}{10.6}{13.1}
& \bootci{-3.0}{-5.3}{-0.7}
& \bootci{+20.2}{11.1}{29.3}
& \bootci{+9.3}{6.9}{11.8}
& \bootci{+13.7}{10.6}{16.9}
& \bootci{\textbf{+25.5}}{23.3}{27.6} \\

\textbf{\oursmodel{} (Qwen 2.5 Omni 7B)}
& \bootci{+26.5}{19.0}{34.0}
& \bootci{+61.5}{54.0}{68.5}
& \bootci{+28.3}{19.7}{37.0}
& \bootci{+13.2}{12.0}{14.5}
& \bootci{-3.4}{-5.9}{-1.0}
& \bootci{+14.6}{5.6}{23.7}
& \bootci{+11.1}{8.6}{13.6}
& \bootci{+15.0}{11.8}{18.3}
& \bootci{\textbf{+20.9}}{18.7}{23.1} \\

\bottomrule
\end{tabular}
}
\caption{\red{
Paired bootstrap analysis of \oursmodel{} improvements over the corresponding base models using 10,000 resamples.
Each cell reports the absolute performance difference ($\Delta$) and its 95\% confidence interval in brackets.}
}
\label{tab:bootstrap_ci}
\end{table*}

\begin{table}[t]
\centering
\small
\resizebox{0.7\columnwidth}{!}{%
\begin{tabular}{lcc}
\toprule
Encoder 
& \makecell{Genre Cls.} 
& \makecell{Key Cls.} \\
\midrule
Qwen2Audio$^\dagger$  & 89.99 & 34.10 \\
MERT & 84.67 & 36.07 \\
\oursmodel{} & 76.00 & 29.28 \\
\bottomrule
\end{tabular}%
}
\vspace{-0.5em}
\caption{\red{
Accuracy comparison between the encoders on two downstream music understanding tasks.
Results marked with $^\dagger$ are taken from~\citet{ghosh2025music}.}
}
\vspace{-12pt}
\label{tab:encoder_ablation_downstream}
\end{table}

\begin{figure*}[t]
\centering
\fbox{
\begin{minipage}{\textwidth}
\footnotesize
\ttfamily

\textbf{[000--032]} \textcolor{descblue}{<Description>}  This music unfolds at a steady 120 BPM in the bright and open key of G major. The overall soundscape is warm and inviting, with a moderate-to-loud presence that creates a sense of gentle momentum. The texture is dominated by a single, continuous melodic instrument that drives the piece forward with a soft, lyrical attack. Its timbre is rich and mellow, suggesting a plucked string instrument like a kora or a harp, playing flowing, arpeggiated patterns…

\textbf{[032]} \textcolor{boundred}{<Boundary>} In perceived mood, emotional arousal decreases noticeably. Musically, The overall loudness remains moderate, with a continuous texture before and after. The harmony modulates from a strong G major center to a strong C major center. Rhythmically, the tempo remains steady. Rhythmically, the rhythmic flow stays steady and moderate. Instrument-wise, Other Instrument stays present with a stable role.

\textbf{[032--154]} \textcolor{descblue}{<Description>} This is a vibrant and rhythmically engaging piece in C major with a tempo of 120 BPM. The overall soundscape is warm and inviting, driven by a continuous and steady melodic core. The music is built around a prominent, bright-sounding plucked string instrument—likely a kora or a similar West African harp …

\textbf{[154]} \textcolor{boundred}{<Boundary>} In perceived mood, emotional arousal increases slightly. Musically, The overall loudness remains moderate, with a continuous texture before and after. The harmony stays stable around a strong C major center. Rhythmically, the tempo remains steady. Rhythmically, the rhythmic flow stays steady and moderate. Instrument-wise, Vocals enters and becomes audible; Other Instrument stays present with a stable role.

\textbf{[154--175]} \textcolor{descblue}{<Description>} This music segment unfolds at a steady 120 BPM in the bright and open key of C major. The overall soundscape is warm and intimate, with a moderate volume that creates a relaxed yet engaging atmosphere. The texture is driven by a continuous melodic line that forms the core of the piece. A female voice enters with a soft, …

\textbf{[175]} \textcolor{boundred}{<Boundary>} In perceived mood, emotional arousal decreases noticeably. Musically, The overall loudness remains moderate, with a continuous texture before and after. The harmony stays stable around a strong C major center. Rhythmically, the rhythmic flow stays steady and moderate, and the tempo undergoes a dramatic shift. Instrument-wise, Vocals drops out of the mix; Other Instrument stays present with a stable role.

\textbf{[175--245]} \textcolor{descblue}{<Description>} This music unfolds at a brisk 150 BPM in the key of C major, creating a vibrant and energetic atmosphere. The overall soundscape is defined by a continuous, driving pulse, with a melodic core …

\end{minipage}
}
\caption{Example of timestamped music caption with \textcolor{descblue}{static} and \textcolor{boundred}{dynamic} captions. (Before rewriting)}
\label{fig:timestamped_caption_before_rewriting}
\end{figure*}

\begin{figure*}[t]
\centering
\fbox{
\begin{minipage}{\textwidth}
\footnotesize
\ttfamily

\textbf{[000--032]} \textcolor{descblue}{<Description>}  This music unfolds at a steady 120 BPM in the bright and open key of G major. The overall soundscape is warm and inviting, with a moderate-to-loud presence that creates a sense of gentle momentum. The texture is dominated by a single, continuous melodic instrument that drives the piece forward with a soft, lyrical attack. Its timbre is rich and mellow, suggesting a plucked string instrument like a kora or a harp, playing flowing, arpeggiated patterns…

\textbf{[032]} \textcolor{boundred}{<Boundary>} In perceived mood, emotional arousal decreases noticeably. Musically, The overall loudness remains moderate, with a continuous texture before and after. The harmony modulates from a strong G major center to a strong C major center. Rhythmically, the tempo remains steady. Rhythmically, the rhythmic flow stays steady and moderate. Instrument-wise, Other Instrument stays present with a stable role.

\textbf{[032--154]} \textcolor{descblue}{<Description>} Continuing the warm momentum, the music now settles into a bright C‑major center at a steady 120 BPM while the overall arousal gently eases. The continuous, steady melodic core persists, driven by the same luminous plucked‑string voice—its clear, resonant timbre shimmering like a West African kora, …

\textbf{[154]} \textcolor{boundred}{<Boundary>} In perceived mood, emotional arousal increases slightly. Musically, The overall loudness remains moderate, with a continuous texture before and after. The harmony stays stable around a strong C major center. Rhythmically, the tempo remains steady. Rhythmically, the rhythmic flow stays steady and moderate. Instrument-wise, Vocals enters and becomes audible; Other Instrument stays present with a stable role.

\textbf{[154--175]} \textcolor{descblue}{<Description>} With the slight rise in arousal, a soft, ethereal female voice slips into the texture, its wordless, breathy phrases hovering above the established melodic current. The bright C‑major backdrop and steady 120 BPM pulse remain unchanged, and the warm, resonant plucked‑string voice that has been threading the piece continues its intricate, finger‑plucked arpeggios…

\textbf{[175]} \textcolor{boundred}{<Boundary>} In perceived mood, emotional arousal decreases noticeably. Musically, The overall loudness remains moderate, with a continuous texture before and after. The harmony stays stable around a strong C major center. Rhythmically, the rhythmic flow stays steady and moderate, and the tempo undergoes a dramatic shift. Instrument-wise, Vocals drops out of the mix; Other Instrument stays present with a stable role.

\textbf{[175--245]} \textcolor{descblue}{<Description>} As the breathy female vocal fades away, the tempo leaps dramatically to a brisk 150 BPM while the harmony stays rooted in C major. The surge in pulse is tempered by a noticeable drop in emotional arousal, so the music feels calmer even as it propels forward. A continuous, driving pulse maintains the moderate volume and steady texture established earlier. The primary voice remains the plucked‑string instrument …

\end{minipage}
}
\caption{Example of timestamped music caption with \textcolor{descblue}{static} and \textcolor{boundred}{dynamic} captions. (After rewriting)}
\label{fig:timestamped_caption_after_rewriting}
\end{figure*}

\begin{table}[t]
\centering
\small
\resizebox{\columnwidth}{!}{%
\begin{tabular}{llrrr}
\toprule
\textbf{Category} & \textbf{Instrument} & \textbf{Count} & \textbf{Rate} & \textbf{Total} \\
\midrule
\multirow{3}{*}{Drums}
    & Cymbals & 30,702 & 73.50\% & \multirow{3}{*}{73.58\%} \\
    & Mallet percussion & 27 & 0.06\% & \\
    & Drums & 8 & 0.02\% & \\
\midrule
Vocals
    & Vocals & 5,070 & 12.14\% & 12.14\% \\
\midrule
\multirow{4}{*}{Plucked strings}
    & Guitar & 3,769 & 9.02\% & \multirow{4}{*}{9.10\%} \\
    & Banjo & 31 & 0.07\% & \\
    & Mandolin & 0 & 0.00\% & \\
    & Ukulele & 0 & 0.00\% & \\
\midrule
\multirow{2}{*}{Brass}
    & Trumpet & 564 & 1.35\% & \multirow{2}{*}{2.33\%} \\
    & Trombone & 408 & 0.98\% & \\
\midrule
\multirow{2}{*}{Bowed strings}
    & Violin & 331 & 0.79\% & \multirow{2}{*}{1.20\%} \\
    & Cello & 170 & 0.41\% & \\
\midrule
\multirow{3}{*}{Winds}
    & Saxophone & 168 & 0.40\% & \multirow{3}{*}{0.68\%} \\
    & Flute & 114 & 0.27\% & \\
    & Clarinet & 1 & 0.00\% & \\
\midrule
\multirow{3}{*}{Keys}
    & Piano & 212 & 0.51\% & \multirow{3}{*}{0.59\%} \\
    & Accordion & 18 & 0.04\% & \\
    & Organ & 16 & 0.04\% & \\
\midrule
Synthesizer
    & Synthesizer & 162 & 0.39\% & 0.39\% \\
\midrule
Bass
    & Bass & 3 & 0.01\% & 0.01\% \\
\midrule
\textbf{Total}
    & -- & \textbf{41,774} & -- & \textbf{100.00\%} \\
\bottomrule
\end{tabular}%
}
\caption{\red{Distribution of added instrument labels grouped into nine instrument categories after cross-validation.}}
\label{tab:instrument_distribution_before}
\end{table}

\begin{table}[t]
\centering
\small
\resizebox{\columnwidth}{!}{%
\begin{tabular}{llrrr}
\toprule
\textbf{Category} & \textbf{Instrument} & \textbf{Count} & \textbf{Rate} & \textbf{Total} \\
\midrule
\multirow{3}{*}{Keys}
    & Piano & 23,987 & 51.34\% & \multirow{3}{*}{51.96\%} \\
    & Organ & 288 & 0.62\% & \\
    & Accordion & 0 & 0.00\% & \\
\midrule
\multirow{3}{*}{Drums}
    & Drums & 14,441 & 30.91\% & \multirow{3}{*}{30.93\%} \\
    & Cymbals & 5 & 0.01\% & \\
    & Mallet percussion & 5 & 0.01\% & \\
\midrule
Synthesizer
    & Synthesizer & 3,310 & 7.08\% & 7.08\% \\
\midrule
Bass
    & Bass & 3,193 & 6.83\% & 6.83\% \\
\midrule
Vocals
    & Voice/vocals & 949 & 2.03\% & 2.03\% \\
\midrule
\multirow{4}{*}{Plucked strings}
    & Guitar & 492 & 1.05\% & \multirow{4}{*}{1.13\%} \\
    & Mandolin & 30 & 0.06\% & \\
    & Banjo & 5 & 0.01\% & \\
    & Ukulele & 2 & 0.00\% & \\
\midrule
\multirow{2}{*}{Bowed strings}
    & Cello & 6 & 0.01\% & \multirow{2}{*}{0.02\%} \\
    & Violin & 5 & 0.01\% & \\
\midrule
\multirow{3}{*}{Winds}
    & Saxophone & 2 & 0.00\% & \multirow{3}{*}{0.01\%} \\
    & Flute & 1 & 0.00\% & \\
    & Clarinet & 0 & 0.00\% & \\
\midrule
\multirow{2}{*}{Brass}
    & Trumpet & 0 & 0.00\% & \multirow{2}{*}{0.00\%} \\
    & Trombone & 0 & 0.00\% & \\
\midrule
\textbf{Total}
    & -- & \textbf{46,721} & -- & \textbf{100.00\%} \\
\bottomrule
\end{tabular}%
}
\caption{\red{Distribution of removed instrument labels grouped into nine instrument categories after cross-validation.}}
\label{tab:instrument_distribution_after}
\end{table}

\begin{table*}[t]
\centering
\renewcommand{\arraystretch}{1.0}
\resizebox{\textwidth}{!}{%
\begin{tabular}{@{}llp{10cm}@{}}
\toprule
\textbf{Category} & \textbf{Feature} & \textbf{Description \& Usage in Pipeline} \\ 
\midrule

\multirow{4}{*}{\textbf{Tonal \& Harmonic}} 
& Key \& Scale & Root note and mode (e.g., C Major); used to identify tonal centers and track modulations across boundaries. \\
& Key Strength & Confidence score ($0.0 \sim 1.0$) of the key prediction; determines the clarity and ambiguity of the harmony. \\
& Chroma & 12-dimensional pitch class profile; cosine distance between segments measures the degree of harmonic palette shifts. \\
& Dissonance & Measure of sensory dissonance; utilized to evaluate whether musical tension grows or resolves during a transition. \\ 
\midrule

\multirow{2}{*}{\textbf{Rhythmic}}
& BPM & Beats per minute; computes absolute changes ($\Delta$BPM) to detect drastic or subtle tempo accelerations/decelerations. \\
& Onset Strength Mean & Average intensity of rhythmic attacks; captures the driving force, busyness, and energy of the rhythmic texture. \\ 
\midrule

\multirow{2}{*}{\textbf{Energy \& Dynamics}}
& Volume & Loudness level used to compute energy shifts and strictly detect silence or instrument entries/exits. \\
& Active Ratio & The proportion of active audio frames; describes the density and textural busyness of the mix and individual stems. \\ 
\midrule

\multirow{1}{*}{\textbf{Timbral \& Semantic}}
& Spectral Centroid & The center of mass of the spectrum (in Hz); serves as a proxy for identifying the brightness or darkness of the timbre. \\

\bottomrule
\end{tabular}%
}
\caption{Summary of extracted musical features utilized for timestamped music caption generation.}
\label{tab:extracted_features}
\end{table*}

\begin{table}[t]
\centering
\small
\setlength{\tabcolsep}{6pt}
\resizebox{\columnwidth}{!}{%
\begin{tabular}{lcccc}
\toprule
\textbf{Hyperparameter}
& \makecell{\textbf{Encoder} \\ \textbf{Pretraining}}
& \makecell{\textbf{Caption} \\ \textbf{Pretraining}}
& \textbf{QA FT}
& \textbf{GRPO} \\
\midrule
LoRA rank & 16 & -- & 64 & 64 \\
LoRA alpha & 32 & -- & 128 & 128 \\
Learning rate & 1e-4 & $5.0{\times}10^{-6}$ & $2.0{\times}10^{-4}$ & $5.0{\times}10^{-6}$ \\
MERT projector learning rate & -- & $2.0{\times}10^{-5}$ & $2.0{\times}10^{-5}$ & $5.0{\times}10^{-7}$ \\
Effective batch size & 512 & 16 & 16 & 32 \\
Training data & 17K & 17K & 40K & 8K \\
Training duration & 1epoch & 1 epoch & 1 epoch & 1000 steps \\
\bottomrule
\end{tabular}%
}
\caption{\red{
Training hyperparameters for each training stage.}
}
\label{tab:training_details}
\end{table}




\begin{figure*}[t]
    \centering

    \begin{subfigure}{0.48\textwidth}
        \centering
        \includegraphics[width=\linewidth]{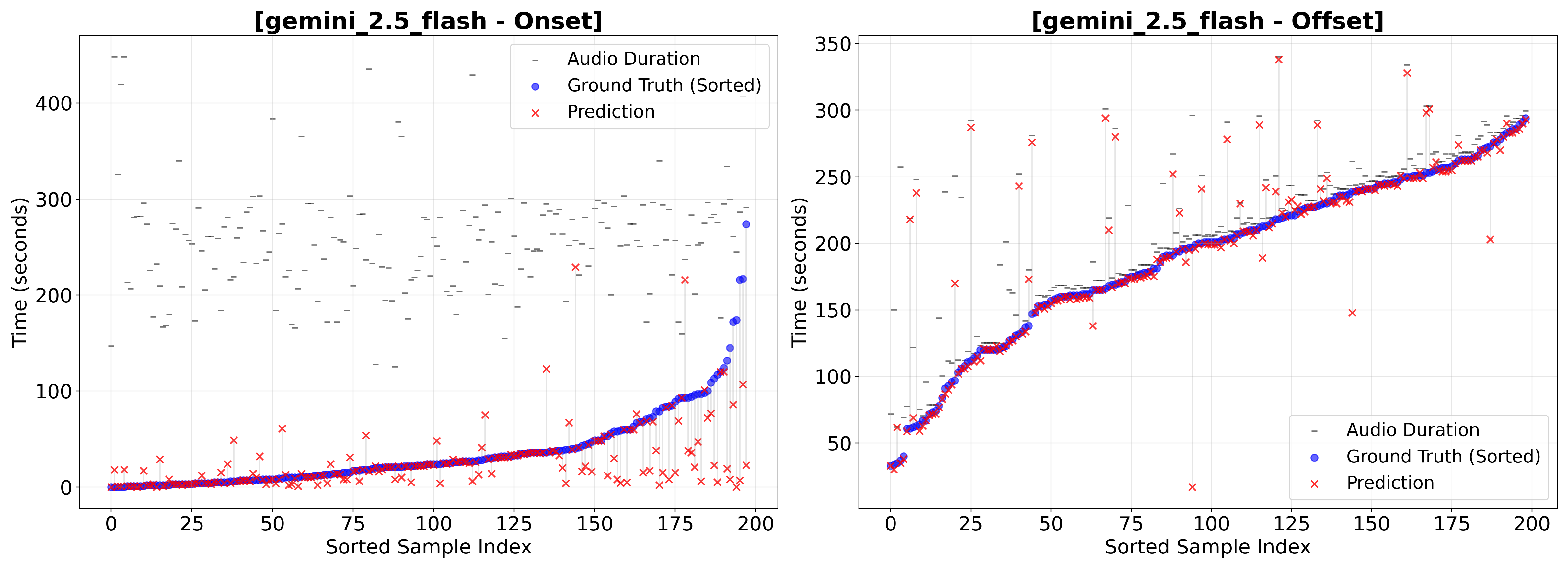}
        \caption{Gemini 2.5 Flash}
        \label{fig:sub1}
    \end{subfigure}
    \hfill
    \begin{subfigure}{0.48\textwidth}
        \centering
        \includegraphics[width=\linewidth]{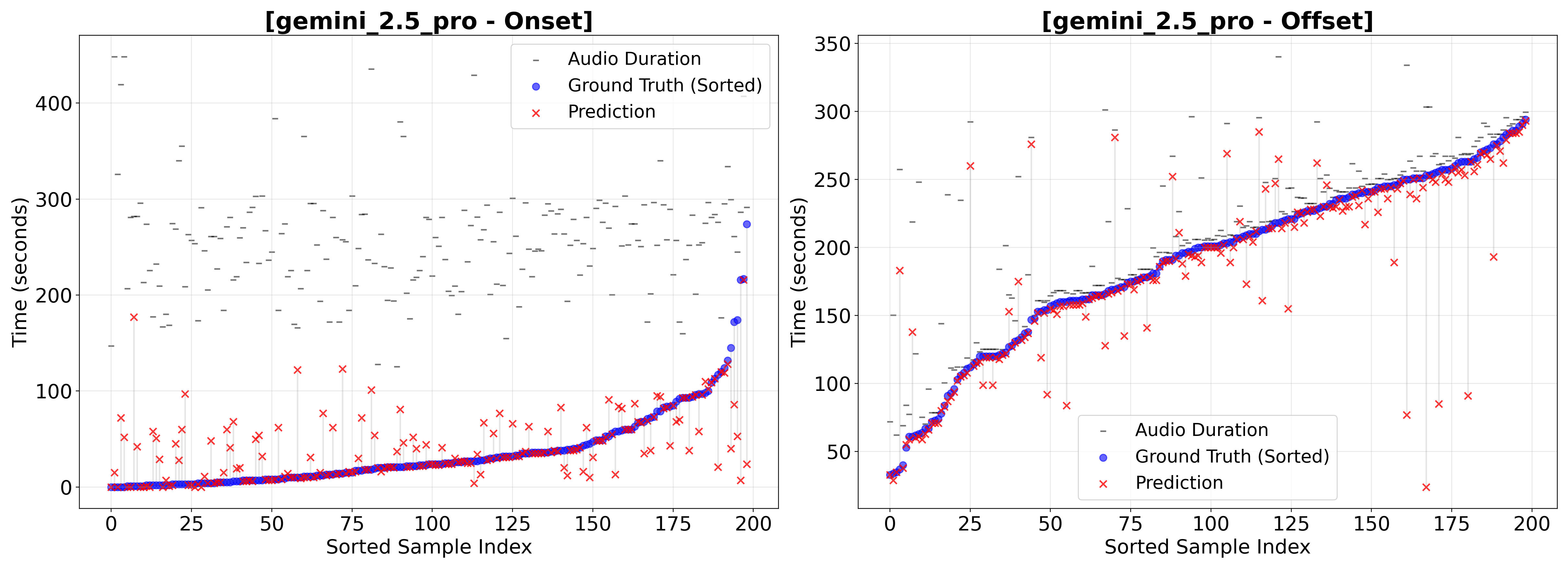}
        \caption{Gemini 2.5 Pro}
        \label{fig:sub2}
    \end{subfigure}

    \vspace{0.5em}

    \begin{subfigure}{0.48\textwidth}
        \centering
        \includegraphics[width=\linewidth]{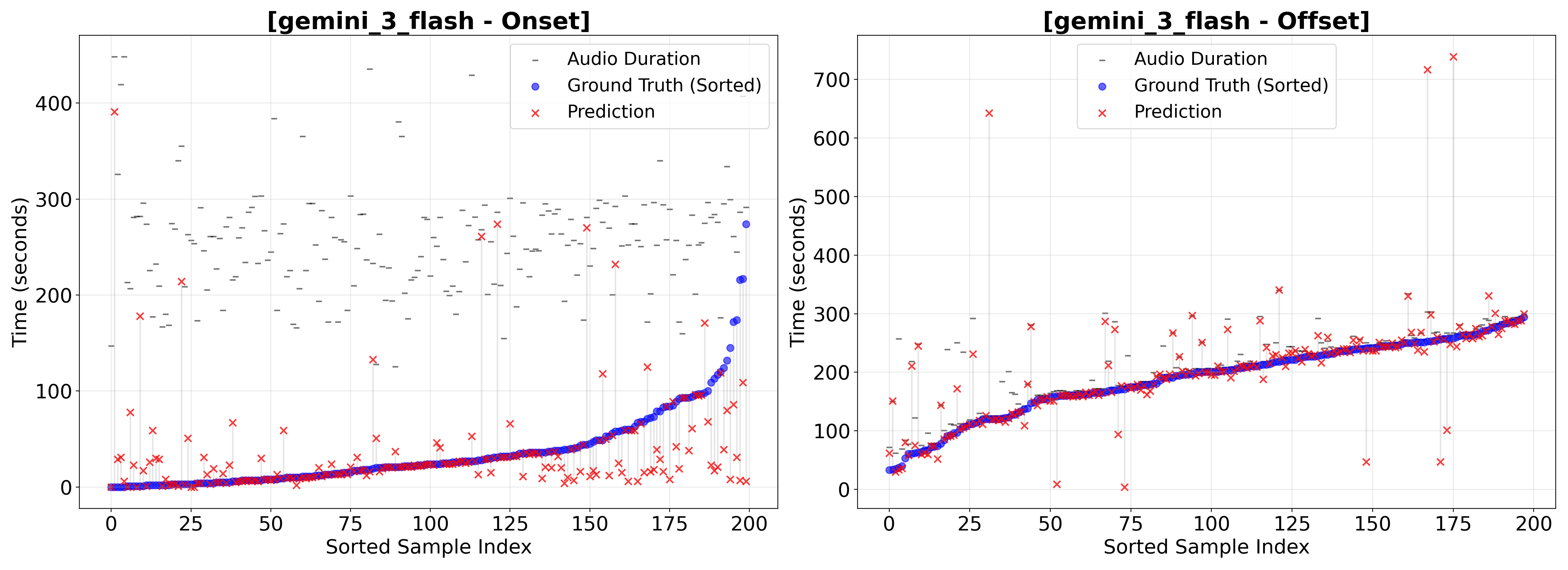}
        \caption{Gemini 3 Flash}
        \label{fig:sub3}
    \end{subfigure}
    \hfill
    \begin{subfigure}{0.48\textwidth}
        \centering
        \includegraphics[width=\linewidth]{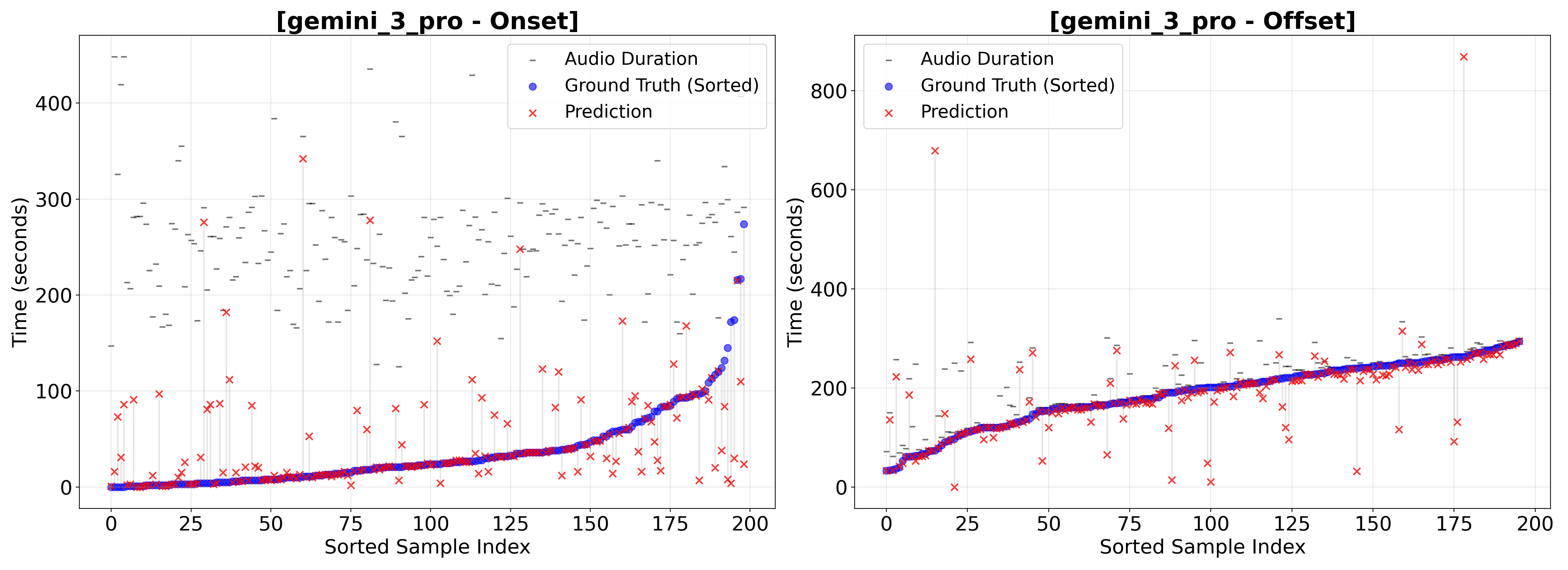}
        \caption{Gemini 3 Pro}
        \label{fig:sub4}
    \end{subfigure}

    \vspace{0.5em}

    \begin{subfigure}{0.48\textwidth}
        \centering
        \includegraphics[width=\linewidth]{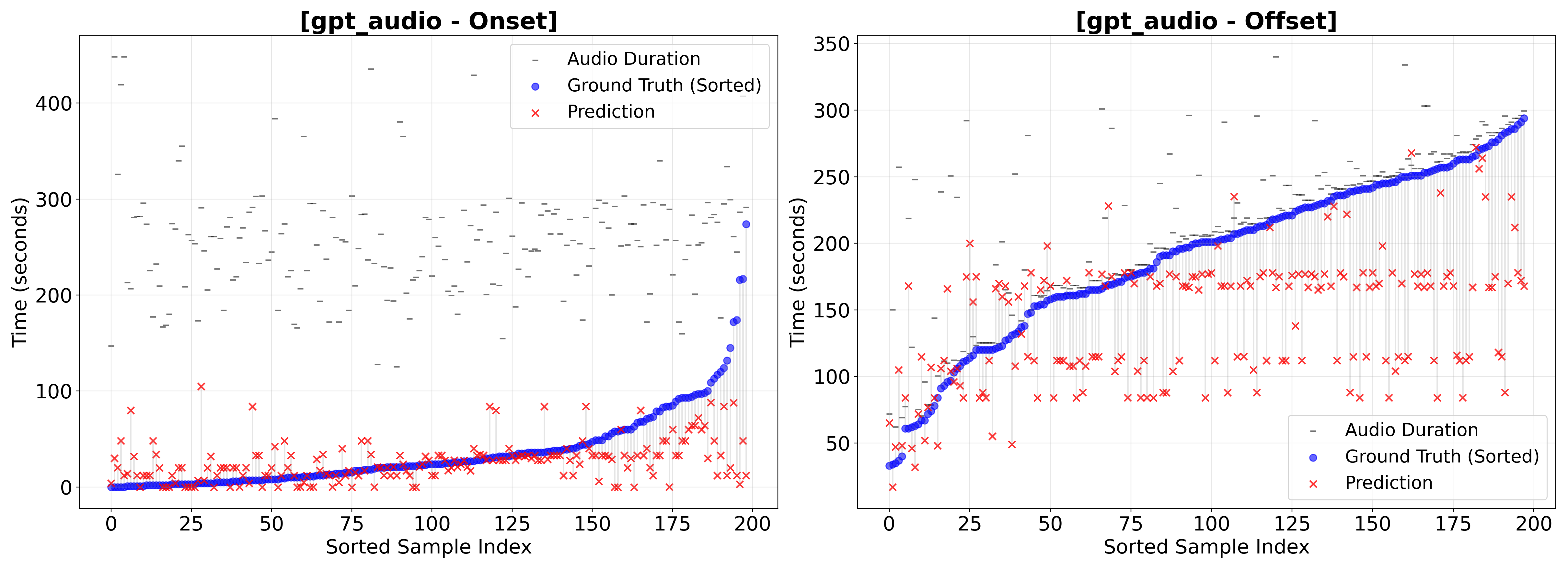}
        \caption{GPT-Audio}
        \label{fig:sub5}
    \end{subfigure}
    \hfill
    \begin{subfigure}{0.48\textwidth}
        \centering
        \includegraphics[width=\linewidth]{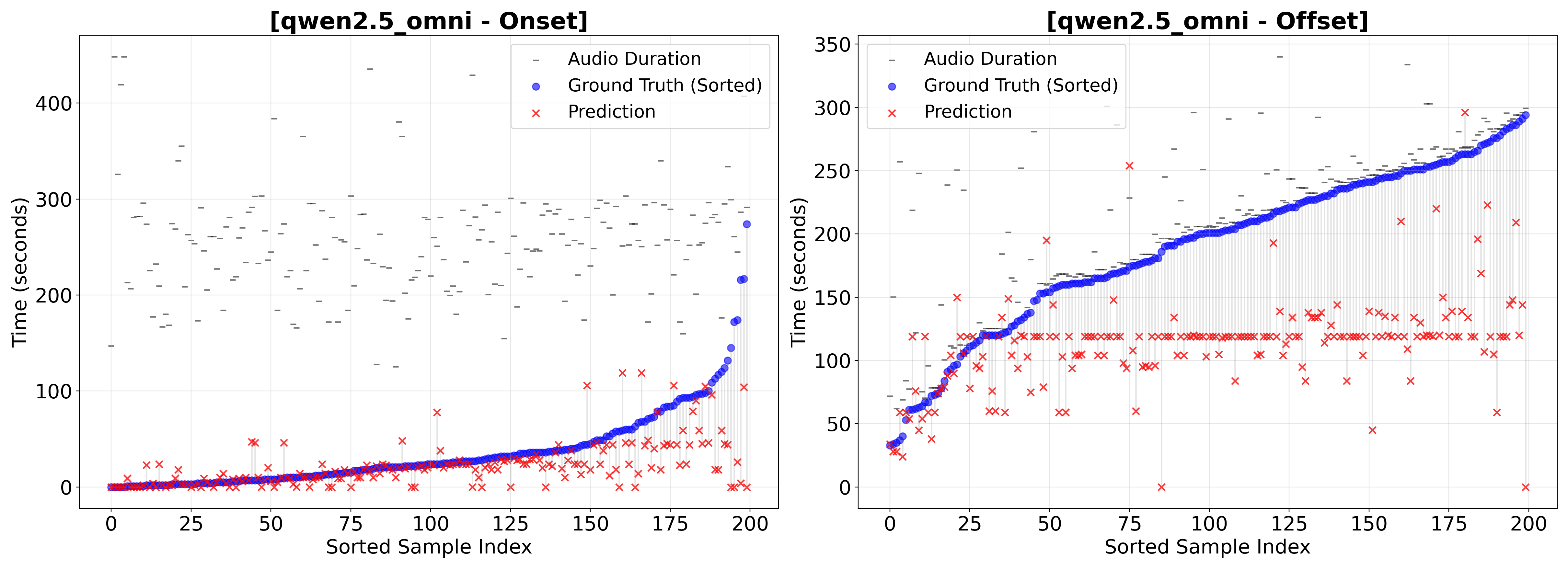}
        \caption{Qwen 2.5 Omni}
        \label{fig:sub6}
    \end{subfigure}

    \vspace{0.5em}

    \begin{subfigure}{0.48\textwidth}
        \centering
        \includegraphics[width=\linewidth]{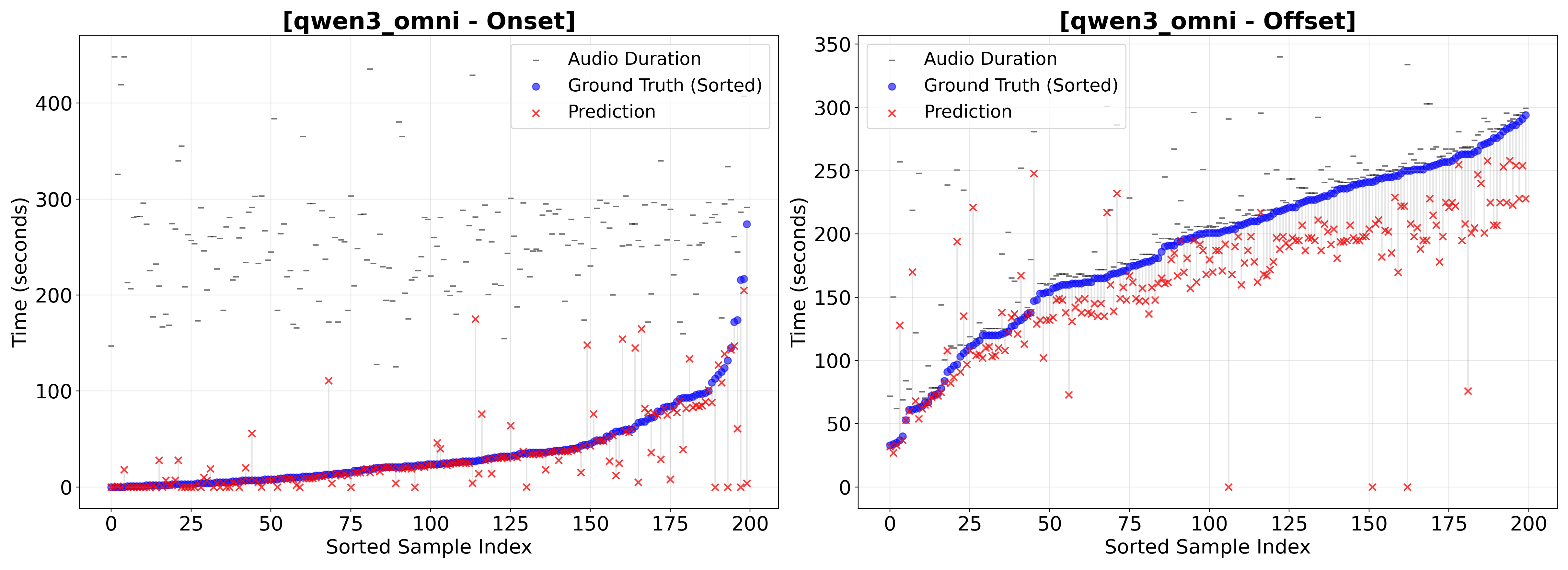}
        \caption{Qwen 3 Omni}
        \label{fig:sub7}
    \end{subfigure}
    \hfill
    \begin{subfigure}{0.48\textwidth}
        \centering
        \includegraphics[width=\linewidth]{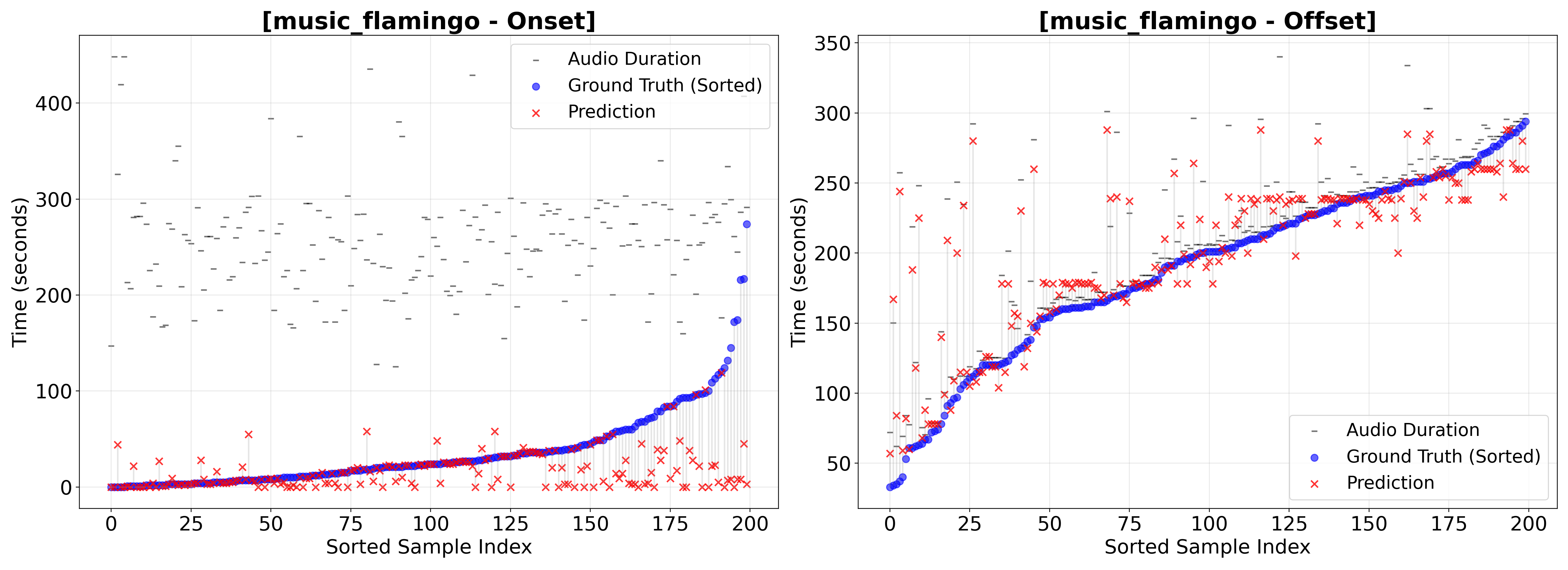}
        \caption{Music Flamingo}
        \label{fig:sub8}
    \end{subfigure}

    \caption{Temporal source grounding (TSG) results across baseline models. \textcolor{blue}{\(\bullet\)} denotes the ground-truth timestamp, and 
\textcolor{red}{\(\times\)} denotes the predicted timestamp.}
    \label{fig:eight_figures}
\end{figure*}

\subsubsection{Experimental Details}
\label{app:experimental_details}

\paragraph{Baseline Models.}
We evaluate representative open-source and closed-source Large Audio-Language Models (LALMs) that support music or general audio understanding.
For open-source baselines, we include Qwen-2.5-Omni~\citep{xu2025qwen2}, Qwen-3-Omni~\citep{xu2025qwen3}, MusicFlamingo~\citep{ghosh2025music}, AudioFlamingo-Next~\citep{ghosh2026audio}, and Phi-4-mm~\citep{abouelenin2025phi}.
For closed-source baselines, we evaluate GPT Audio and GPT Audio 1.5~\citep{hurst2024gpt}, Gemini 2.5 Pro/Flash~\citep{comanici2025gemini}, and Gemini 3 Pro/Flash~\citep{google2025gemini3}.
All baseline models are evaluated in a zero-shot setting to assess their inherent temporal grounding ability without task-specific adaptation.

\paragraph{Inference Protocol.}
For each benchmark example, we provide the model with the full music track and the corresponding task-specific instruction.
No task-specific demonstrations or in-context examples are provided.
All models are evaluated with temperature set to 0 to ensure deterministic inference.
For multiple-choice tasks, we extract the selected option from the model response.
For timestamp and interval prediction tasks, we parse temporal expressions from the generated text and treat invalid formats as incorrect.

\begin{table}[t]
\centering
\small
\resizebox{\columnwidth}{!}{%
\begin{tabular}{lllll}
\toprule
\textbf{Stage}
& \textbf{Training Step}
& \textbf{Time}
& \textbf{Hardware}
& \textbf{Peak Memory} \\
\midrule

Stage 2
& Caption Pretraining
& 3 hours
& 8$\times$H200 GPUs
& 113 GB/GPU \\

Stage 3
& QA LoRA Fine-tuning
& 3 hours
& 8$\times$H200 GPUs
& 93 GB/GPU \\

Stage 4
& GRPO Training
& 4 hours
& 8$\times$H200 GPUs
& 49 GB/GPU \\
\bottomrule
\end{tabular}%
}
\caption{\red{
Training time and memory requirements for LLM training stages.}
}
\label{tab:training_efficiency}
\end{table}
\begin{table}[t]
\centering
\small
\resizebox{\columnwidth}{!}{%
\begin{tabular}{ll}
\toprule
\textbf{Component}
& \textbf{Inference Overhead} \\
\midrule
Additional Resident VRAM
& $\sim$0.69 GiB \\

Additional \oursmodel{} Tokens
& 3.33 tokens/s \\

Maximum \oursmodel{} Tokens
& 1,000 for 300-s audio \\

Additional KV Cache
& Up to $\sim$55 MiB per 300-s request \\

Decode Overhead
& Relatively small for short QA answers \\
\bottomrule
\end{tabular}%
}
\caption{\red{
Inference overhead introduced by \oursmodel{}.}
}
\label{tab:inference_overhead}
\end{table}

\paragraph{Training Details.}
\red{We train our model on 8$\times$H200 GPUs using the AdamW optimizer~\citep{loshchilov2017decoupled} with a cosine learning rate scheduler~\citep{loshchilov2016sgdr}.
We set maximum token budget to 2000 for Qwen tokens (6.66 tokens/sec) and 1000 for \oursmodel{} tokens (3.33 tokens/sec). We apply dynamic sampling only to \oursmodel{} tokens, keeping uniform sampling for Qwen tokens.
Detailed hyperparameters for each training stage are summarized in Table~\ref{tab:training_details}.
For GRPO training, we sample 8K examples from the QA training set, consisting of 4K examples each from TSG and MTR.
We set the number of rollouts to 8, with a temperature of 1.0, top-p of 0.95, and a clipping ratio of 0.2.
We additionally report training time and peak memory usage for each main training stage (Table~\ref{tab:training_efficiency}), and the additional inference overhead (Table~\ref{tab:inference_overhead}).}

\begin{figure*}[t]
\centering
\fbox{
\begin{minipage}{\textwidth}
\footnotesize
\ttfamily

What best describes the musical change at 133s?

A. …

B. …

C. …

D. …

You need to listen around T-10 - T+10, then fill ‘answer’ column with

(If there is correct answer) correct option → A/B/C/D

(If there’s no correct answer) your own correct sentence 
(description) → There’s vocal entry and …

\end{minipage}
}
\caption{Human-assisted validation instruction for LTR and TAD.}
\label{fig:human_annotation_instruction_type2}
\end{figure*}
\begin{figure*}[t]
\centering
\fbox{
\begin{minipage}{\textwidth}
\footnotesize
\ttfamily

The following descriptions (X, Y, Z) represent three distinct musical transitions from the same track.

Descriptions:

(X) [200s] …

(Y) [15s] …

(Z) [160s] …

Question: Which of the following options represents the correct chronological order of these transitions as they appear in the track?

A. X → Y → Z

B. X → Z → Y

C. Y → X → Z

D. Y → Z → X  << already known

E. Z → X → Y

F. Z → Y → X

You need to listen X, Y, and Z (around T-10 - T+10) to check whether the descriptions are correct.

If they are correct, please write “Correct”.

If there is a wrong description, please type wrong transition with reason. (ex. X: There is no guitar entry)

If there are multiple wrong descriptions, please use comma separated values. 

(ex. X: There is no guitar entry, Y: There is no key change)

\end{minipage}
}
\caption{Human-assisted validation instruction for GTO.}
\label{fig:human_annotation_instruction_type4}
\end{figure*}
\begin{figure*}[t]
\centering
\fbox{
\begin{minipage}{\textwidth}
\footnotesize
\ttfamily

Question

When is(are) the highest arousal duration(s) in this audio?

Please answer with duration in seconds (ex. [72-110])

If there are multiple timestamps for highest/lowest duration, you can list multiple durations.

Ex) [55-110], [130-160]

Arousal is the level of energy and intensity.

\end{minipage}
}
\caption{Human-assisted validation instruction for MTR.}
\label{fig:human_annotation_instruction_type5}
\end{figure*}



\end{document}